\documentclass[sigconf]{acmart}
\AtBeginDocument{%
  }

\copyrightyear{2026}
\acmYear{2026}
\setcopyright{cc}
\setcctype{by-nc-nd}
\acmConference[WWW '26]{Proceedings of the ACM Web Conference 2026}{April 13--17, 2026}{Dubai, United Arab Emirates}
\acmBooktitle{Proceedings of the ACM Web Conference 2026 (WWW '26), April 13--17, 2026, Dubai, United Arab Emirates}
\acmPrice{}
\acmDOI{10.1145/3774904.3792589}
\acmISBN{979-8-4007-2307-0/2026/04}

\usepackage{comment}
\usepackage{enumitem}
\usepackage{multirow}

\usepackage{algorithm}
\usepackage{algpseudocode}  %
\usepackage{amsmath}        %
\acmSubmissionID{rfp3241}
\begin{document}

\title{GraphRAG-R1: Graph Retrieval-Augmented Generation with Process-Constrained Reinforcement Learning}

\newcommand{\methodname}{GraphRAG-R1}
\newcommand{\RLName}{Rollout Retrieval Enhanced GRPO}
\newcommand{\rlname}{rollout retrieval enhanced GRPO}
\newcommand{\RewardName}{Process-Constrained Reward}
\newcommand{\rewardname}{process-constrained reward}

\author{Chuanyue Yu}
\orcid{0009-0004-7818-9808}
\affiliation{
  \institution{Nankai University}
   \city{Binhai New Area}
  \state{Tianjin}
  \country{China}
}
\email{yuchuanyue@mail.nankai.edu.cn}

\author{Kuo Zhao}
\orcid{0009-0003-9033-2112}
\affiliation{
  \institution{Nankai University}
   \city{Binhai New Area}
  \state{Tianjin}
  \country{China}
}
\email{zhaokuo@mail.nankai.edu.cn}

\author{Yuhan Li}
\orcid{0000-0003-1324-5819}
\affiliation{
  \institution{HKUST Guangzhou}
   \city{Guangzhou}
   \state{Guangdong} 
  \country{China}
}
\email{yuhanli98@gmail.com}

\author{Heng Chang}
\orcid{0000-0002-4978-8041}
\authornote{Project co-leader.}
\affiliation{
  \institution{Huawei Technologies Ltd}
  \city{Haidian District}
  \state{Beijing}
  \country{China}
}
\email{changh.heng@gmail.com}

\author{Mingjian Feng}
\orcid{0009-0007-0345-762X}
\affiliation{
  \institution{Nankai University}
  \city{Binhai New Area}
  \state{Tianjin}
  \country{China}
}
\email{fengmingjian@mail.nankai.edu.cn}

\author{Xiangzhe Jiang}
\orcid{0009-0007-4243-9029}
\affiliation{
  \institution{Nankai University}
  \city{Binhai New Area}
  \state{Tianjin}
  \country{China}
}
\email{jiangxiangzhe@mail.nankai.edu.cn}

\author{Yufei Sun}
\orcid{0009-0001-9778-9575}
\affiliation{
  \institution{Nankai University}
   \city{Binhai New Area}
  \state{Tianjin}
  \country{China}
}
\email{yufei_sun@sina.com}

\author{Jia Li}
\orcid{0000-0002-6362-4385}
\affiliation{
  \institution{HKUST Guangzhou}
  \city{Guangzhou}
  \state{Guangdong} 
  \country{China}
}
\email{jialee@hkust-gz.edu.cn}

\author{Yuzhi Zhang}
\orcid{0000-0002-6729-925X}
\authornotemark[1]
\affiliation{
  \institution{Nankai University}
   \city{Binhai New Area}
  \state{Tianjin}
  \country{China}
}
\email{zyz@nankai.edu.cn}

\author{Qingyun Sun}
\orcid{0000-0003-1930-3848}
\affiliation{
  \institution{Beihang University}
  \city{Haidian District}
  \state{Beijing}
  \country{China}
}
\email{sunqy@buaa.edu.cn}

\author{Jianxin Li}
\orcid{0000-0001-5152-0055}
\affiliation{
  \institution{Beihang University}
  \city{Haidian District}
  \state{Beijing}
  \country{China}
}
\email{lijx@buaa.edu.cn}

\author{Ziwei Zhang}
\orcid{0000-0003-2451-843X}
\authornotemark[1]
\authornote{Corresponding author.}
\affiliation{
  \institution{Beihang University}
  \city{Haidian District}
  \state{Beijing}
  \country{China}
}
\email{zwzhang@buaa.edu.cn}

\renewcommand{\shortauthors}{Chuanyue Yu et al.}

\begin{abstract}
Graph Retrieval-Augmented Generation (GraphRAG) has shown great effectiveness in enhancing the reasoning abilities of Large Language Models (LLMs) by leveraging graph structures for knowledge representation and modeling complex real-world relationships. However, existing GraphRAG methods still face significant bottlenecks when handling complex problems that require multi-hop reasoning, as  %
their query and retrieval phases are largely based on pre-defined heuristics and do not fully utilize the reasoning potentials of LLMs.  
To address this problem, we propose \methodname, an adaptive GraphRAG framework by training LLMs with process-constrained outcome-based reinforcement learning (RL) to enhance the multi-hop reasoning ability. Our method can decompose complex problems, autonomously invoke retrieval tools to acquire necessary information, and perform effective reasoning.
Specifically, we utilize a modified version of Group Relative Policy Optimization (GRPO) that supports rollout-with-thinking capability to train the model. Next, we design two process-constrained reward functions. To handle the shallow retrieval problem, we design a Progressive Retrieval Attenuation (PRA) reward to encourage essential retrievals. Then, to handle the over-thinking problem, we design a Cost-Aware F1 (CAF) reward to balance the model performance with computational costs. 
We further design a phase-dependent training strategy, containing three training stages corresponding to cold start and these two rewards. These stages empower GraphRAG with format following, behavior shaping, and smartness optimization abilities, respectively.
Lastly, our method adopts a hybrid graph-textual retrieval to improve the reasoning capacity.
Extensive experimental results demonstrate that \methodname~significantly boosts LLM capabilities in solving complex reasoning problems compared to state-of-the-art GraphRAG methods on both in-domain and out-of-domain datasets. Furthermore, our framework can be flexibly integrated with various existing retrieval methods, consistently delivering performance improvements.

\end{abstract}



\maketitle
\newcommand\webconfavailabilityurl{https://doi.org/10.5281/zenodo.18349146}
\ifdefempty{\webconfavailabilityurl}{}{
\begingroup\small\noindent\raggedright\textbf{Resource Availability:}
The source code has been made publicly available at \url{https://doi.org/10.5281/zenodo.18349146} and can also be accessed via the project repository: \url{https://github.com/ycygit/GraphRAG-R1}. The model weights are available at \url{https://doi.org/10.57967/hf/7629} and \url{https://huggingface.co/yuchuanyue/GraphRAG-R1}.
\endgroup
}
\section{Introduction}
\begin{figure*}
  
  \includegraphics[width=\textwidth,page=1]{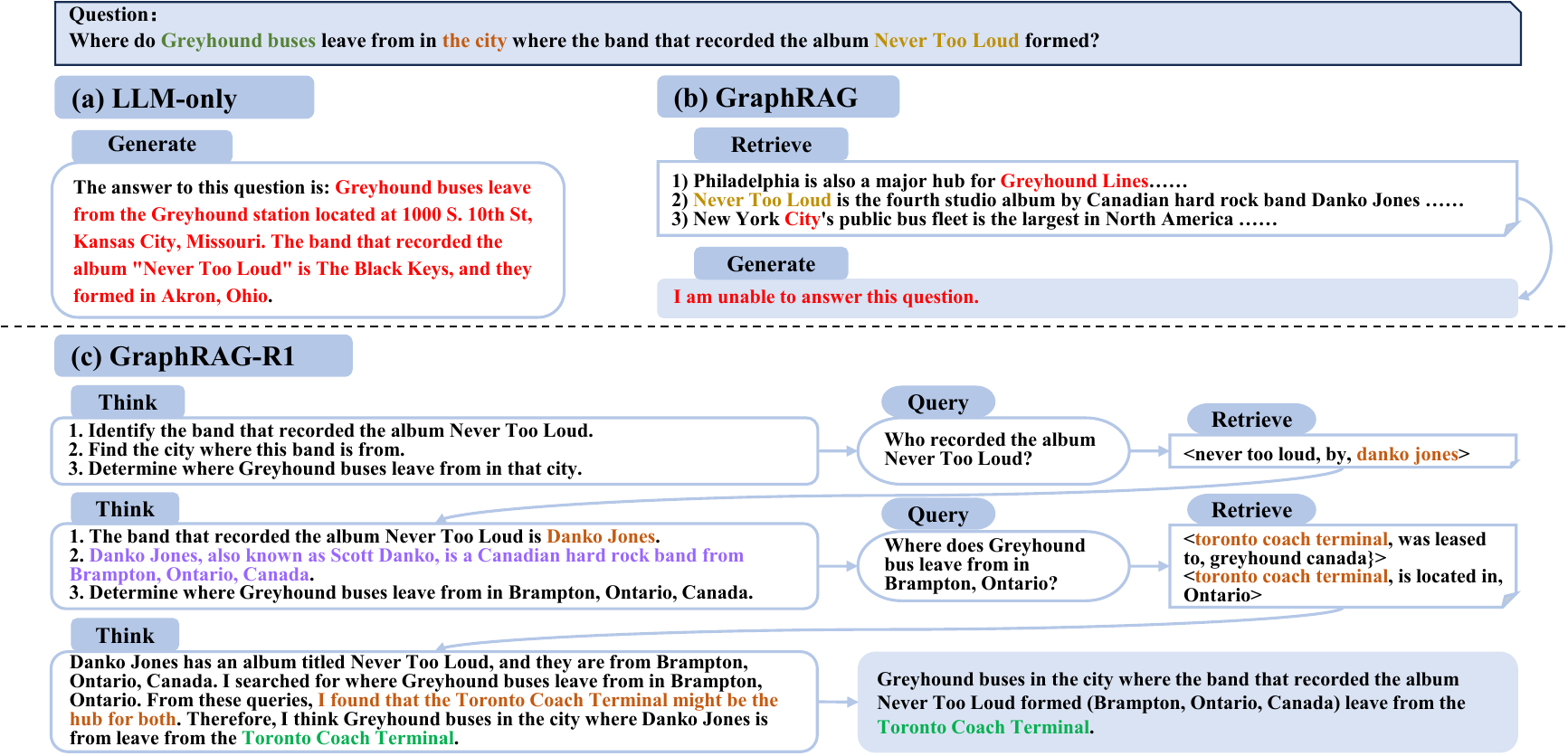}
  \vspace{-0.8cm}
  \caption{An example of comparing using only LLMs, GraphRAG, and \methodname~in answering complex problems: (a) The LLM (Qwen2.5-7B) directly produces an output, but the answer is incorrect, (b) HippoRAG2~\cite{gutierrez2025hipporag2} enhances the LLM (Qwen2.5-7B) by retrieving external knowledge, but fail to generate correct outputs for complex problem,  (c) our method successfully decomposes the problem and autonomously invokes retrieval methods, producing the correct output.}
  \Description{This figure provides a comparative schematic illustrating the performance of three distinct approaches in handling the same complex reasoning problem, visually demonstrating the superiority of the proposed GraphRAG-R1 framework. The schematic is divided into three juxtaposed sub-figures:
(a) Using only the base Large Language Model (LLM, Qwen2.5-7B): The model directly generates an answer, which is incorrect. This reveals the limitations of LLMs regarding knowledge scope and multi-step reasoning.
(b) Using an existing Graph Retrieval-Augmented Generation method (HippoRAG2) to augment the same LLM: While this method introduces external knowledge retrieval, it still fails to produce the correct answer for this complex problem. This highlights the bottlenecks of traditional GraphRAG methods in flexible problem decomposition and multi-hop reasoning.
(c) Applying the proposed GraphRAG-R1 method: The model successfully decomposes the complex problem into sub-steps, autonomously and on-demand invokes external retrieval tools to acquire necessary information, and finally integrates the reasoning to produce the correct answer. Through this comparison, the figure clearly validates the effectiveness of GraphRAG-R1 in enhancing the LLM's capability to solve complex problems.}
  \label{fig:example}
\vspace{-0.3cm}
\end{figure*}

Large Language Models (LLMs) have attracted ever-growing attention in the past few years due to their outstanding ability in answering complex questions and generating human-quality text~\cite{vaswani2017attention,wang2024current}.
Despite remarkable progress, LLMs still exhibit significant limitations when faced with complex tasks that require integrating vast and diverse knowledge domains or require deep reasoning and multi-step logical inference~\cite{talmor2019commonsenseqa,petroni2021kilt}. 
As shown in Figure~\ref{fig:example}(a), LLMs tend to provide an incorrect response when the question
requires knowledge beyond their scope.  
Graph Retrieval-Augmented Generation (GraphRAG)~\cite{xiang2025use,zhangsurvey,pengsurvey,hansurvey,zhu2024structugraphrag,li2025g}, which utilizes knowledge graphs for knowledge representation, enables effective modeling of complex real-world relationships, such as causality, hierarchy, and dependency~\cite{micro_graphrag,pengsurvey}.

However, existing GraphRAG methods still face significant performance bottlenecks when handling complex problems that require multi-hop reasoning. The main bottleneck is that existing methods rely on pre-defined or heuristic rules during query processing and retrieval, which are too rigid and inflexible. Concretely, during query processing, the semantic or lexical similarity search~\cite{hansurvey,gao2023retrieval,fan2024survey} remains the most widely adopted. For semantic similarity search, queries with high complexity are averaged out during vectorization, making it difficult to gather accurate information for each distinct semantic element. On the other hand, lexical similarity search captures only the superficial meaning of the query, easily losing implicit relationships and reasoning logic. Therefore, as shown in Figure~\ref{fig:example}(b), existing GraphRAG methods are unable to answer the query, since the retrieval results are incomplete or incorrect when queries are overly complex.
Besides, during the graph retrieval phase, current methods employ a combination of graph traversal and subgraph retrieval for path reasoning, which are rather rigid~\cite{tog,tog2,jimenez2024hipporag,guo2024lightrag,li2024graphreader}. For instance, ToG~\cite{tog,tog2}, a representative GraphRAG method, starts from a node in the graph and traverses the surrounding nodes based on breadth-first search strategy with pruning. While this strategy might work for short paths, it struggles with long-path problems, which require capturing logically sound but distant or even disconnected relationships. 
This inflexibility often leads to broken reasoning chains and failed retrievals, making it unsuitable for multi-hop problems with complex reasoning logic.
As a result, the existing GraphRAG methods fail to fully leverage the LLM's deep reasoning potential.%

In this paper, we try to further enhance the multi-hop reasoning ability of GraphRAG.
Reinforcement learning (RL) has recently been shown to be effective in endowing LLMs with powerful reasoning abilities. For example, in the case of DeepSeek-R1~\cite{guo2025deepseek}, it has been discovered that large-scale reinforcement learning can endow models with deep thinking capabilities~\cite{xie2025logic,jaech2024openai,team2025kimi}, even without any supervised fine-tuning. This effectiveness may result from RL's use of reward signals to encourage exploration, refining dynamic reasoning strategies from static knowledge~\cite{ouyang2022training} beyond static reasoning patterns~\cite{asai2024self,zhang2025corag,xiong2025mcts,deeprag}. However, how to utilize RL for GraphRAG training remains unexplored in the literature. The major technical obstacle lies in how to design proper reward functions so that the reasoning process of LLMs within GraphRAG can be properly constrained, avoiding potential problems such as shallow retrieval~\cite{helwe2021reasoning,deng2025boosting} and over-thinking~\cite{sui2025stop,chen2024not,cuesta2025large}. 

To address this issue, we propose \methodname, an adaptive GraphRAG framework with deep reasoning capabilities through process-constrained outcome-based RL. As shown in Figure~\ref{fig:example}(c), our method can decompose complex multi-hop problems by reasoning and autonomous invocation of retrieval tools. Specifically, we employ a modified version of Group Relative Policy Optimization (GRPO) that supports rollout-with-thinking mechanism~\cite{song2025r1searcher} to leverage external retrieval tools on demand. Next, we design two process-constrained reward functions, Progressive Retrieval Attenuation (PRA) and Cost-Aware F1 (CAF), aiming to inject prior knowledge into the LLM's training process.
In detail, PRA aims to dynamically adjust the reward for retrieval actions. It encourages necessary retrievals
, preventing the model from shallow retrieval. 
As the reasoning deepens, PRA progressively attenuates retrieval rewards, 
guiding the model to 
avoid unnecessary retrieval overhead. 
On the other hand, CAF evaluates the final answer correctness using the F1 score while also accounting for the computational resources consumed during the reasoning and retrieval process. It encourages the model to achieve correct results more efficiently to prevent over-thinking. These two rewards collectively balance effectiveness and computational cost. We further design a three-stage phase-dependent training strategy. In the first stage, we focus on the answer format to alleviate cold start problems. In the second and third stages, we focus on improving the answer quality using the aforementioned two rewards. 
Lastly, our method employs a hybrid graph-textual retrieval.
By leveraging graph structures to precisely capture entity relationships while preserving the rich semantic details of the original text, our method provides more comprehensive and integrated knowledge for complex reasoning.

Experiments show that \methodname~significantly outperforms existing GraphRAG methods in solving complex reasoning problems, across both in-domain and out-of-domain datasets. Additionally, our framework offers flexible integration with various retrieval methods, consistently enhancing their performance.

Our main contributions are summarized as follows:
\begin{itemize}[leftmargin=0.5cm]
\item We propose \methodname, a pioneering framework that utilizes GRPO that supports rollout-with-thinking to enhance the reasoning ability of GraphRAG. %
\item We propose PRA and CAF, two tailored reward designs to guide the training process of RL and handle the shallow retrieval and over-thinking problems. We also propose a phase-dependent training strategy and a hybrid graph-textual retrieval. 

\item Experiments demonstrate that our method exhibits state-of-the-art performance on both in-domain and out-of-domain datasets, and can be flexibly integrated with various existing retrieval methods with consistently improved performance.
\end{itemize}

\section{Related Work}
\subsection{Graph Retrieval-Augmented Generation}
Graph Retrieval-Augmented Generation (GraphRAG) offers a novel paradigm to enhance the reasoning abilities of LLMs 
by explicitly modeling entity relationships through graph structure such as knowledge graphs~\cite{pengsurvey,hansurvey}. %
GraphRAG has been widely adopted across multiple domains. For instance, GraphRAG based on knowledge graph~\cite{pan2024unifying} incorporates techniques like subgraph retrieval and Graph Neural Networks (GNNs)~\cite{he2024grtriever} to enhance the accuracy and explainability of LLMs in complex knowledge reasoning question answering. GraphRAG based on document graphs~\cite{sonawane2014graph} models semantic relationships between documents and internal document structures, effectively supporting long-document summarization and cross-document information integration~\cite{sarthi2024raptor}. Furthermore, GraphRAG methods for specialized domains are increasingly trending. Examples include leveraging social graphs~\cite{jiang2023social,zeng2024large} to analyze user relationships for precision recommendations and constructing medical or disease graphs~\cite{yang2024kg,li2024dalk} that integrate multi-source heterogeneous data to enable personalized treatment plan generation.

However, existing GraphRAG methods still face significant performance bottlenecks when facing complex reasoning problems.
For example, ToG~\cite{tog,tog2} introduces reasoning steps using breadth-first search (BFS) with pruning approach, but it heavily relies on external LLMs and thus lacks flexibility. 
G-Retriever~\cite{he2024grtriever} adopts K-Nearest Neighbors-based retrieval with limited efficacy in multi-hop scenarios. 
Moreover, existing approaches suffer from shallow retrieval depth, since they often depend on predefined fixed paths or hop counts. 
HippoRAG~\cite{jimenez2024hipporag,gutierrez2025hipporag2} 
only supports single-hop retrieval. 
KGP~\cite{wang2024kgp} and LightRAG~\cite{guo2024lightrag} are typically confined to 2-3 hops, exhibiting performance degradation when handling problems that require deep-dependence mining. 
In contrast, our work improves the reasoning ability of GraphRAG for complex problems using process-constrained RL, aiming to 
effectively resolves the issues of rigid retrieval strategies and complex reasoning shortcomings in existing GraphRAG approaches.

\subsection{RL-Enhanced Retrieval Scheme}
Recently, reinforcement learning (RL) has become increasingly important for complex reasoning~\cite{xie2025logic,yang2024dollmreasoning,ouyang2022training}. %
For example, DeepSeek-R1~\cite{guo2025deepseek} shows that large-scale RL can endow LLMs with deep thinking capabilities. 
Some pioneering works have also adopted RL to enhance RAG~\cite{jin2025searchr1,song2025r1searcher}. For example, R1-Searcher~\cite{song2025r1searcher} and Search-R1~\cite{jin2025searchr1} achieve co-training of retrieval and generation, effectively improving the model's ability to solve complex problems. However, a systematic exploration of RL schemes within the GraphRAG domain remains open. Technically speaking, directly extending existing RL methods into GraphRAG faces several challenges. Firstly, the existing methods usually employ a single outcome-oriented reward function, e.g., based solely on the correctness of final answer. However, this design becomes too simple for GraphRAG, 
since a complex problem may involve complex retrieval calls
of the external knowledge graph. 
More critically, existing methods have often encountered reward hacking in RL training, potentially resulting in shallow retrieval and over-thinking in GraphRAG. To tackle these issues, we propose \methodname~with tailored reward functions and training strategy to non-trivially explore the potential of RL in enhancing GraphRAG.

\section{Method}
\begin{figure*}
  \includegraphics[width=\textwidth,page=1]{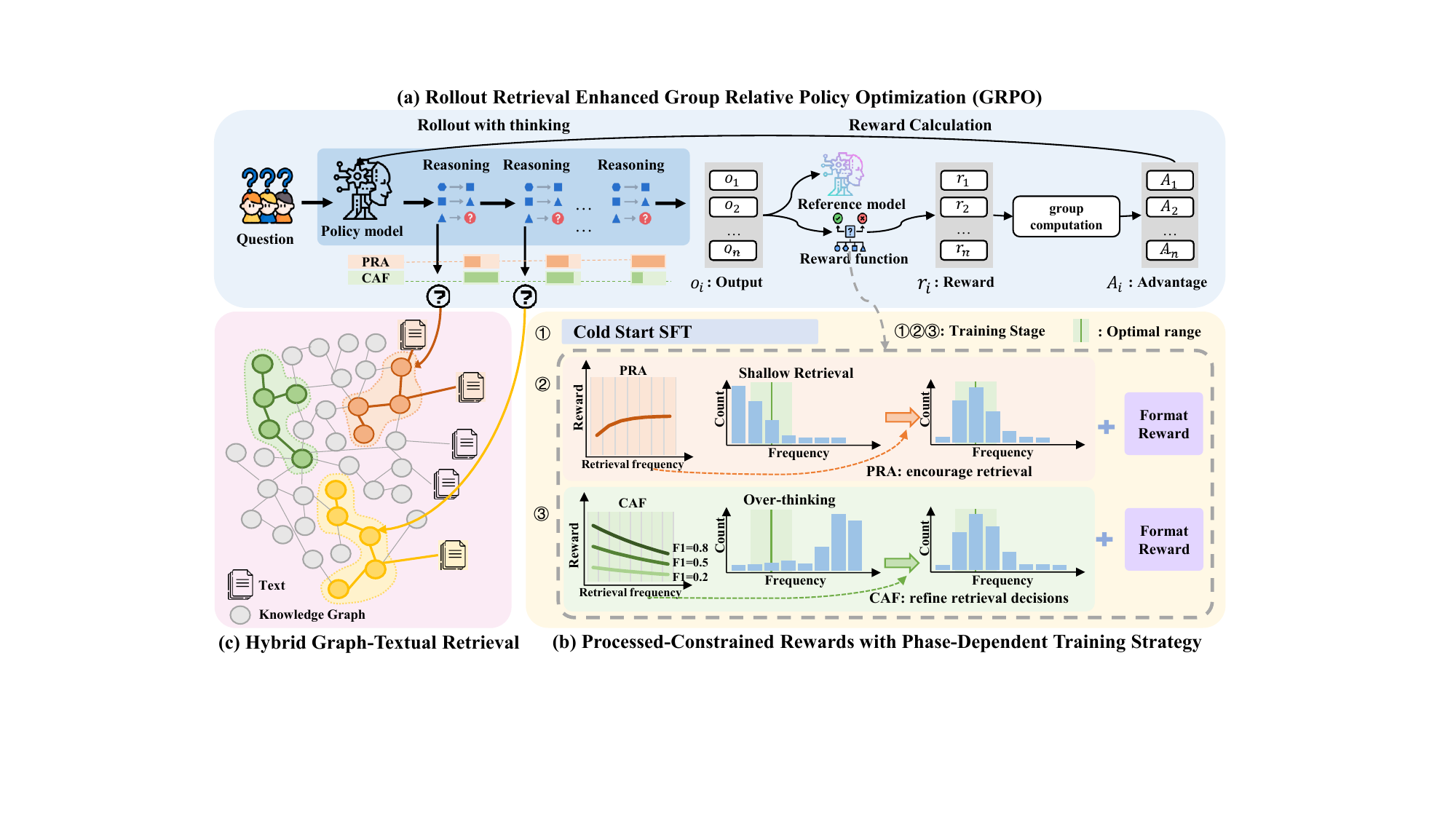}
  \vspace{-0.7cm}
  \caption{An overview of \methodname~: (a) the \rlname~as the training strategy of LLM, (b) \rewardname~designs, containing PRA and CAF rewards and the phase-dependent training strategy, (c) the hybrid graph-textual retrieval, which is more informative than text fragments. }
 \Description{This figure presents a comprehensive architectural overview of the proposed GraphRAG-R1 framework, illustrating its three core, interconnected components:
(a) Rollout Retrieval Enhanced GRPO as the LLM Training Strategy: This component depicts the reinforcement learning (RL) training paradigm, specifically a modified Group Relative Policy Optimization (GRPO) algorithm. It visually represents how the LLM policy is trained via process-constrained outcome-based RL, enabling the model to autonomously invoke retrieval tools during its reasoning "rollout" or thought process.
(b) Process-Constrained Reward Designs: This section details the design of the reward functions that guide and constrain the model's behavior during RL training. It highlights the two novel rewards and illustrates the phase-dependent training strategy that applies them sequentially (cold start, behavior shaping, smartness optimization) to effectively shape the model's reasoning and retrieval patterns.
(c) Hybrid Graph-Textual Retrieval Module: This part contrasts the proposed retrieval mechanism with traditional text-only retrieval. It visually demonstrates how combining structured graph triplets (capturing precise entity relationships) with unstructured text snippets (preserving rich semantic context) provides more comprehensive and integrated knowledge, thereby supplying higher-quality information for the LLM's complex reasoning tasks.
}
  \label{fig:framework}
  \vspace{-0.3cm}
\end{figure*}

In this section, we present our proposed method. First, we introduce the \rlname~algorithm. Second, we detail the designs of \rewardname~, which includes PRA and CAF, together with the phase-dependent training strategy. 
Finally, we present our hybrid graph-textual retrieval design. %
An overall framework of our method is shown in Figure~\ref{fig:framework}.

\subsection{\RLName}
In this subsection, we introduce the details of our training method, which is a modified version of GRPO that supports the roll-out-with-thinking capability~\cite{song2025r1searcher}.

Our goal is to optimize the LLM using RL to generate high-quality outputs augmented by an external retrieval tool $\mathcal{R}$ that assists the reasoning generation process. Specifically, denote the policy of RL as $\pi_{\theta}(\cdot)$ parametrized by $\theta$. Let $q$ represent input queries drawn from the training dataset $\mathcal{D}$, $o$ represent the generated outputs, and $r(\cdot)$ be the reward function. We formalize the objective as follows:
\begin{equation}
\begin{aligned}
&\max_{\pi_{\theta}} \mathbb{E}_{q \sim \mathcal{D}, o \sim \pi_{\theta}(\cdot \mid q;\mathcal{R})} 
\big[ r(q,o) - 
\beta \cdot \mathbb{D}_{\text{KL}}(\pi_\theta  \|  \pi_{\text{ref}}) 
\big] ,
\end{aligned}
\end{equation}
where $\pi_{\text{ref}}(\cdot)$ is the reference LLM, $\mathbb{D}_{\text{KL}}(\cdot)$ is the KL-divergence measure, and $\beta$ is a hyperparameter that controls the strength of the KL-divergence regularization. More specifically, our method builds on the GRPO algorithm~\cite{guo2025deepseek}, which uses the average reward of multiple sampled outputs as a baseline. During training, for each input query $q$, GRPO samples a group of $G$ responses ${o_1, o_2,..., o_G}$ using the old policy $\pi_{\theta_{\text{old}}}$. The policy model is then optimized by maximizing the following objective function:
\begin{equation}\label{eq:GRPO}
\begin{aligned}
&\mathcal{J}_{\text{GRPO}}(\theta) = \mathbb{E}_{\substack{q \sim D \\ \{o_i\}_{i=1}^G \sim \pi_{\theta_{\text{old}}}} }
\bigg[ 
 \frac{1}{G} \sum_{i=1}^{G} \min \bigg( 
\frac{\pi_\theta(o_{i}\mid q)}{\pi_{\theta_{\text{old}}}(o_{i} \mid q)} \cdot A_{i}, \\
& \text{clip}\left( \frac{\pi_\theta(o_{i} \mid q)}{\pi_{\theta_{\text{old}}}(o_{i} \mid q )}, 1-\epsilon, 1+\epsilon \right) \cdot A_{i} 
\bigg) 
 - \beta \cdot \mathbb{D}_{\text{KL}}( \pi_\theta  \|  \pi_{\text{ref}} )
\bigg],
\end{aligned}
\end{equation}
where the group size $G$ determines the number of responses generated per query, $\theta_{\text{old}}$ denotes the previous policy parameters, $\epsilon$ is the clipping range hyperparameter to constrain policy updates, and $A_{i}$ represents the advantage, computed based on the relative rewards of outputs within each group as follows:
\begin{equation}\label{eq:advantage}
A_{i} = \frac{r_i - \mu_r}{\sigma_r},   
\end{equation}
where $r_i$ is the reward for individual output, $\mu_r$ means the average reward of all $G$ outputs for the same query $q$, and $\sigma_r$ is the group reward standard deviation. %

To better adapt GRPO to our GraphRAG scenario, we adopt two key modifications proposed by R1-Searcher~\cite{song2025r1searcher}: Rollout-with-Thinking and Retrieval-Masked Loss. The core objective is to enable the model to autonomously leverage external retrieval tools on demand during inference and receive rewards for generating correct answers. %

\subsubsection{Rollout-with-Thinking} 
The model is guided to utilize the external retrieval tool during generation by using two special tokens \texttt{<|begin\_of\_query|>...<|end\_of\_query|>} to indicate the call to the search tool. 
When \texttt{<|end\_of\_query|>} appears, it signifies that the retrieval query has been fully generated. 
At this point, model generation is paused, the external retrieval tool is called, and the retrieved content is marked with \texttt{<|begin\_of\_documents|> ... <|end\_of\_documents|>}. 
This content then guides the model's next step of reasoning. 
This method ensures retrieval is seamlessly integrated into the reasoning process.

\subsubsection{Retrieval-Masked Loss}
In the original GRPO, token-level loss is computed throughout the rollout sequence. In the GraphRAG scenario, the rollout sequence consists of tokens generated by the LLM and tokens retrieved from external channels. 
Applying the same optimization to retrieved tokens and LLM-generated tokens may negatively impact the training process. Therefore, we utilize special tokens \texttt{<|begin\_of\_documents|>} and \texttt{<|end\_of\_documents|>} and their enclosed content is masked during training, ensuring the policy gradient objective is calculated only on LLM-generated tokens. 

\subsection{\RewardName~Design}
In this subsection, we introduce our reward functions to constrain the retrieval process, enabling effective and efficient reasoning. 

RL methods have recently attracted considerable attention to improve the reasoning abilities of LLMs. However, if simple reward functions are employed, such as relying solely on final answer correctness, existing methods easily face a technical problem known as ``Reward Hacking''~\cite{gao2023scaling}, i.e., the model exploits certain loopholes in the reward to gain high scores rather than genuinely improving reasoning capabilities. %
To address this potential issue, our reward design injects controlled human prior knowledge while preserving the fundamental reward framework. Concretely, we introduce the following three rewards: format reward, PRA, and CAF, as follows. %

\subsubsection{Format reward}

This reward function is designed to steer the model toward strictly adhering to the specified output format, ensuring that it learns to correctly invoke external retrieval and generate the final answer. The model should place the final response entirely within the \texttt{<answer>...</answer>} tags and ensure that the entire output is fully readable and free of uninterpretable artifacts. When invoking retrieval, the model should encapsulate retrieval queries within the \texttt{<|begin\_of\_query|>...<|end\_of\_query|>} tags with retrieved information contained within \texttt{<|begin\_of\_documents|> ... <|end\_of\_documents|>}. The model should explicitly invoke the retrieval mechanism through this query process rather than directly generating content without retrieval. 

The format reward is defined as follows:
\begin{equation}\label{eq:formatreward}
  R_{\text{format}} = 
  \begin{cases} 
  0.5, & \text{correct} \\
  0,   & \text{incorrect}.
  \end{cases}
  \end{equation}

\subsubsection{Progressive Retrieval Attenuation (PRA)}
The most straightforward design to encourage GraphRAG models to invoke retrieval calls is to adopt a single retrieval reward. However, this strategy tends to cause the model to invoke the external retrieval only once and cause the \textit{shallow retrieve} problem, as some complex multi-hop problems require multiple retrieval calls. On the other hand, one can sum up rewards for multiple retrieves, but it may trigger unnecessary retrieval calls. To circumvent these issues, we propose Progressive Retrieval Attenuation (PRA), specifically designed to provide tailored rewards for each invocation while incorporating a decay factor. With each additional retrieval call, the reward undergoes exponential decay relative to the base reward, concurrently accumulating rewards from the preceding invocations. Formally, the retrieval reward is defined as follows:
\begin{equation}\label{eq:PRA}
   R_n = 
   \begin{cases}
      R_{0}, & n=1 \\
      R_{n-1} + R_{0} \times k^{n-1}, & n>1 
   \end{cases},
   \end{equation}
where $R_0$ is the base reward, $n$ is the number of retrieval calls, and $0 < k < 1$ is the decay factor. 
Through this dynamically adjusted reward weighting, the retrieval frequency of the model can be precisely controlled within the desired threshold ranges. Using the properties of geometric series, it is easy to see that:
\begin{equation}
    R_0 \leq R_n < \frac{1}{1-k} R_0, \; \forall n.
\end{equation}
When $k=0$, PRA degenerates into a single retrieval reward. When $k=1$, it degenerates into the sum of rewards. The final reward is $R_{\text{retrieval}} = R_N$, where $N$ is the actual number of retrieval calls. 

Through this reward function, we can effectively restrict the number of retrieval calls, effectively preventing shallow retrieval. For different data characteristics, the hyperparameter $k$ can be flexibly adjusted to better steer the model's learning process.

\subsubsection{Cost-Aware F1 (CAF)}\label{sec:CAF}
Although the PRA reward can regulate the retrieval frequency of the model within a reasonable range, this mechanism alone still relies on prior knowledge and could not fully activate the reasoning capabilities of LLMs. For example, the LLM may adopt unnecessarily numerous retrievals, leading to the ``over-thinking'' problem and greatly waste the computational resources.   
An ideal GraphRAG model should deliver high-quality responses with precisely needed retrieval operations, which requires three core competencies: accurately assessing the problem difficulty, intelligently decomposing complex queries, and dynamically optimizing retrieval strategies.  

To this end, we design the CAF reward, which achieves the above objective through another decay coefficient. Formally, the CAF reward is defined as:
\begin{equation}\label{eq:CAF}
   R_{\text{CAF}}= F1 \times a \times  e^{-b \times N} ,
\end{equation} 
where $F1$ is the F1 score of the generated answers, $a$ and $b$ are hyperparameters to control the coefficients in the regulatory function, and $N$ denotes the total number of retrieval operations. The CAF reward ensures that responses achieving identical F1 score receive higher rewards when using fewer retrievals, thereby incentivizing the model to proactively recognize problem complexity and refine retrieval decisions. 
Consequently, CAF reward establishes a balance among reasoning depth, retrieval cost, and generation quality, effectively mitigating suboptimal behaviors of over-thinking.

\subsubsection{Phase-Dependent Training Strategy}
Despite the effectiveness of these reward functions, simultaneously optimizing them induces technical challenges. %
To address this challenge, we propose a phase-dependent training strategy to avoid the potential conflicts between objectives. Concretely, we split the training into three sequential stages, detailed as follows.

Specifically, in the first stage, we utilize another LLM to generate data following our desired format. Then, we utilize supervised fine-tuning (SFT) to train the LLM with the generated data. This training process can enable the model to follow our desired format by addressing the cold start problem and stabilizing the training process, laying the foundation for the following training process.

In the second stage, we aim to shape the behavior of the model, synergistically combining the format reward and the PRA reward. This design recognizes standardized tool-calling protocols and retrieval behavior as essential prerequisites for high-quality reasoning. By learning with two objectives, the model can efficiently learn how and when to invoke the retrieval tools with structured output patterns. The decay mechanism of PRA dynamically regulates retrieval behavior through a high initial reward, driving essential information retrieval, while its exponential decay characteristic suppresses redundant operations. 

In the last stage, we further optimize the overall smartness of the model, adopting CAF as the primary reward objective while retaining format rewards to prevent forgetting issues. As discussed in Section~\ref{sec:CAF}, CAF establishes a trade-off between maximizing answer correctness and constraining retrieval counts. This incentivizes models to adaptively adjust retrieval depth based on query complexity, suppressing non-essential retrievals for simple queries while expanding information acquisition for complex multi-hop problems, leading to effective and efficient reasoning. 

In summary, our phase-dependent training strategy resolves potential gradient conflicts, %
improving training stability and significantly boosting model performance. 

\subsection{Hybrid Graph-Textual Retrieval}
Compared to conventional RAG, GraphRAG exhibits superior relational modeling capabilities using graph structure. %
We integrate this advantage into both the training and generation phases of our framework. %
Specifically, we retrieve both structured triplets and relevant texts, replacing text fragments as commonly retrieved in existing RAGs and GraphRAGs with hybrid graph-textual retrieval.

By employing hybrid retrieval, our method has several advantages. 
Compared to lengthy text fragments, the compact nature of graph can drastically reduce token consumption, improving processing efficiency. 
This structured representation
allows the model to focus directly on essential entity-relationships while inherently filtering out the redundant noise. 
Therefore, the integration of structured knowledge significantly enhances the quality and efficiency of generated answers, as reasoning operates on a cleaner and more relevant information foundation. We empirically verify the hybrid graph-textual retrieval in Section~\ref{sec:exp:hybrid}.

\begin{table*}
  \caption{The results of different methods for multi-hop question answering. The first three datasets are in-domain, and PopQA is out-of-domain (i.e., unseen during training). The best results are in \textbf{bold} and the second-best results are \underline{underlined}.}
  \vspace{-0.4cm}
  \label{tab:main}
  \begin{tabular}{lcccccccccccc}
    \toprule
     \multirow{2}{*}{Method}      & \multicolumn{3}{c} {HotpotQA}    & \multicolumn{3}{c}{MuSiQue}    & \multicolumn{3}{c}{2Wiki}  & \multicolumn{3}{c}{PopQA}\\

              \cmidrule[1pt](lr){2-4}\cmidrule[1pt](lr){5-7}\cmidrule[1pt](lr){8-10}\cmidrule[1pt](lr){11-13}\
                            & $\mathrm{F1}$     & $\mathrm{ACC_{L}}$        & $\mathrm{SBERT}$       & $\mathrm{F1}$          & $\mathrm{ACC_{L}}$          & $\mathrm{SBERT}$     & $\mathrm{F1}$        & $\mathrm{ACC_{L}}$          & $\mathrm{SBERT}$   & $\mathrm{F1}$  & $\mathrm{ACC_{L}}$        & $\mathrm{SBERT}$\\
 \midrule
       \text{Vanilla LLM}          &   3.08 &  30.00 &    44.32 &  2.01 &   15.50   &   43.14&  4.55&   29.50  &  44.37 &   10.46   &  25.90 & 50.07\\
          \text{Naive RAG} & 24.24  &   49.00   &   58.73  &  8.99   & 19.00   &  48.97  &   13.14  &  25.00   &  52.56  &   23.72    &   \underline{58.10}   &     58.60    \\
       \text{Prompt Engineering} &   22.22  &    41.00 &    58.67  & 7.72  &    \underline{25.50}  &   49.91 &  \underline{17.54}&   36.00  &  \underline{54.44} & 13.37    & 27.50 &51.61\\

       \text{SFT}                   &   12.10 & 49.50  &   51.76  &    8.29  &   21.00  &      48.90  & 11.40 &   \underline{48.50}  &  49.46  & 10.45     & 54.10 &50.56\\

    \cmidrule[1pt](lr){1-13}

       \text{KGP}                &  10.73   &   21.00    &  50.92  &   4.61  &   11.00  &  46.70   & 10.16   & 18.00    &  50.27   &  21.01 & 50.00 & 56.10  \\
        \text{ToG}                &  11.44 &  21.50   &   50.48 &  5.02 &   8.00     &  47.05 &  14.45 &     21.50   &  53.11  &  \underline{29.21} &  47.00  &  \underline{62.94}   \\
         \text{LightRAG}        &   7.96  & 18.50  &   49.15 &   3.09 & 6.00   &  44.99  &   7.58  &  10.00   &  47.95 &  14.55&  34.30  &   53.07   \\

        \text{PropRAG}    &  25.47 &  \textbf{55.00}  &   59.76  &  10.11 &    21.50   &  \underline{50.77} &   17.17 &   35.00  &  53.85 &  23.29  &  57.30  & 58.22      \\
         \text{G-retriever}    &   14.81  &   21.50   &   54.89  &   3.38 &   5.00  &  47.47 &  15.12   &    19.00  &   52.77  &  2.14  &    6.80  &   44.63   \\
        \text{HippoRAG2}        &  \underline{27.52}  &  \underline{53.00}  & \underline{61.02}    &  8.90  &  23.50   & 50.19  & 12.32     & 29.00   & 51.92   &  23.99&  57.46 &   58.16     \\
         \text{R1-Searcher}     &26.82   &  35.00  & 59.15 &   \underline{12.35} &  14.50  &  50.32 &  15.89 &  22.50 & 51.51 &  23.24 & 43.20 &  56.80  \\
    \cmidrule[1pt](lr){1-13}

         \methodname        &  \textbf{38.00} &  52.00   & \textbf{66.66}   & \textbf{20.06}  &  \textbf{26.50}    & \textbf{57.49} &  \textbf{32.24}  &  \textbf{49.50}   & \textbf{63.56}       &   \textbf{35.04}    &    \textbf{60.00}  & \textbf{65.31}\\

    \bottomrule
  \end{tabular}
   \vspace{-0.3cm}
\end{table*}

\section{Experiments}
In this section, we conduct experiments to evaluate \methodname, in order to answer the following research questions: 
\begin{itemize}[leftmargin=0.4cm]
    \item \textbf{RQ1:} Does \methodname~ effectively improve the performance of existing LLMs for both in-domain and out-of-domain datasets?
    \item \textbf{RQ2:} Do all components, including PRA and CAF rewards, and phase-dependent training, contribute positively to our method? %
    \item \textbf{RQ3:}  Is \methodname~ flexible enough to be seamlessly integrated with other retrievers? 
\end{itemize}

\subsection{Experimental Setup}
\subsubsection{Dataset}
To comprehensively evaluate the \methodname~ framework, we follow~\cite{jimenez2024hipporag,wang2025proprag} and select four widely used multi-hop QA benchmark datasets: HotpotQA~\cite{yang2018hotpotqa}, 2Wiki~\cite{ho2020constructing}, MuSiQue~\cite{trivedi2021musique}, and PopQA~\cite{mallen2023not}. 
For the training process, we utilize 80\% of the training data from HotpotQA, 2Wiki, and MuSiQue. 
The remaining 20\% of these datasets are employed as the in-domain testing set. 
Meanwhile, we utilize PopQA as an out-of-domain testing set, aiming to compare the generalization capabilities of different models. More details are provided in Appendix~\ref{sec:dataset}.

\subsubsection{Metrics}
Following the common literature, we employ F1 Score~\cite{wang2024kgp}, SBERT Similarity (SBERT)~\cite{thakurbeir}, and LLM-as-Judge Accuracy ($\mathrm{ACC_{L}}$) ~\cite{song2025r1searcher} as evaluation metrics. 
The F1 Score measures the lexical overlap between generated answers and ground-truth references. SBERT uses the cosine similarity between sentence embeddings generated by the SBERT model to measure semantic similarity. $\mathrm{ACC_{L}}$ leverages an LLM (specifically, Qwen3-Turbo) as an evaluator.

\subsubsection{Baselines}
We compare 7 representative methods:
\begin{itemize}[leftmargin=0.5cm]
        \item \textbf{KGP}~\cite{wang2024kgp}: It utilizes fine-tuned LLMs to explore neighboring nodes and aggregate information from a KG to produce answers.
        \item \textbf{ToG}~\cite{tog}: It dynamically expands and evaluates the most probable inference path in the KG to perform reasoning.
        \item \textbf{LightRAG}~\cite{guo2024lightrag}: It enhances standard RAG by integrating graph-based text indexing with a dual-level retrieval strategy.
        \item \textbf{PropRAG}~\cite{wang2025proprag}: It optimizes the reasoning chain via propositional representation and beam search.
        \item \textbf{G-Retriever}~\cite{he2024grtriever}: It integrates LLM and graph neural network (GNN) with RAG for conversational graph QA.
        \item \textbf{HippoRAG2}~\cite{gutierrez2025hipporag2}: It uses personalized PageRank for deep passage integration inspired by the hippocampal indexing theory.
        \item \textbf{R1-Searcher}~\cite{song2025r1searcher}: It enhances the capability of LLMs to invoke RAG for retrieval through a two-stage RL framework.
\end{itemize}
We also compare with the following self-constructed baselines:
\begin{itemize}[leftmargin=0.5cm]
        \item \textbf{Vanilla LLM}: It uses the raw backbone LLM for inference. 
        \item \textbf{Naïve RAG}: It adopts a single-hop retrieval from the vector database for generation.
        \item \textbf{Prompt Engineering}: It adopts GraphRAG via structural prompts.
        \item \textbf{SFT}: It adopts GraphRAG via SFT on the KGQA data.
\end{itemize}

\subsubsection{Implementation Details}
We utilize Qwen2.5-7B as the default backbone LLM model. For baselines that require external closed-source LLMs, we use Qwen3-Turbo. 
We use HippoRAG2 as the external retriever for LLM's rollout retrieval.

For training, in the first stage, we generate training data using ERNIE-3.5-128K~\cite{wang2021ernie}.  %
In the second stage, we set the base reward for PRA as $R_0 = 0.5$ and its decay factor as $k=0.5$.
In the third stage, we set the hyper-parameter for CAF as $a=2$ and $b=0.1$. %
Throughout three stages, LoRA fine-tuning~\cite{hu2021lora} is employed with $r=16$ and $\text{lora}\_{\text{alpha}}=32$. 
The sampling temperature is set to $1.0$ with a maximum retrieval count of $8$, the learning rate is $5e-6$, and the number of epochs for SFT is $1$.

\subsection{Main Results}
To answer \textbf{Q1}, we present the performance comparison between \methodname~ and baselines across four benchmarks in Table~\ref{tab:main}.%
We can draw the following three observations:

\methodname~ achieves consistent improvements across all datasets. Specifically, we achieve the best results for evaluation metrics F1 and SBERT in all datasets, and achieve three best results and one third-best result for $\mathrm{ACC_{L}}$.  
This fully validates the effectiveness of the \methodname~ framework in enhancing LLMs' capability to handle multi-hop reasoning problems. Particularly, %
it outperforms the strongest baseline in F1 score by up to 38.08\% on HotpotQA, 62.43\% on MuSiQue, and 83.81\% on 2Wiki, respectively. \methodname~ gets more improvement on a more difficult multi-hop dataset.

\methodname~achieves leading performance on both in-domain datasets
and the out-of-domain dataset PopQA. Concretely, it outperforms the best baseline by up to 19.96\% in terms of F1 on PopQA, indicating that our proposed method possesses strong generalization ability and can effectively handle unseen complex questions. %

From the results of self-constructed baselines, existing techniques such as naive RAG, prompt engineering, and SFT can indeed improve the ability of vanilla LLM on complex question answering tasks. However, our method, which adopts RL-based training, is clearly more effective, usually outperforming all these non-RL approaches by a large margin. The consistent performance improvement validates that RL-based methods hold more potential in enhancing GraphRAG and the reasoning ability of LLMs in general.

\begin{table}
\setlength{\tabcolsep}{2pt}
  \caption{The ablation study results on PRA and CAF rewards. \#Calls denotes the number of retrieval calls.}
  \vspace{-0.4cm}
  \label{tab:pra-caf}
  \resizebox{0.48\textwidth}{!}{
  \begin{tabular}{lcccccccc}
    \toprule
    \multirow{2}{*}{Method}     & \multicolumn{2}{c} {HotpotQA}    & \multicolumn{2}{c}{MuSiQue}    & \multicolumn{2}{c}{2Wiki} & \multicolumn{2}{c}{PopQA}\\

              \cmidrule[1pt](lr){2-3}\cmidrule[1pt](lr){4-5}\cmidrule[1pt](lr){6-7}\cmidrule[1pt](lr){8-9}\
                             & $\mathrm{F1}$                      & $\mathrm{\#Calls}$                                    & $\mathrm{F1}$                     & $\mathrm{\#Calls}$                                    & $\mathrm{F1}$                     & $\mathrm{\#Calls}$     & $\mathrm{F1}$                     & $\mathrm{\#Calls}$ \\
    \midrule
        \text{\methodname}        &   \textbf{38.00}        &    1.94       &  \textbf{20.06}      &    2.36    &   3\textbf{2.24}         &  2.05        &         \textbf{35.04}      & 1.49\\

    \text{ - PRA}     &    \underline{33.92}         &   1.77      &  14.48       &    2.22      &   \underline{31.55}        &  1.75       &      \underline{31.40}       &1.32\\
    \text{ - CAF}     &   28.36         &    2.33     &   \underline{19.82}      &      2.80   &      22.16       &      2.36       &      26.98      & 2.05\\
    \text{ - PRA \& CAF} &     33.34       &    1.83     &    15.32    &    2.26     &    28.94         &       1.94     &    30.06       & 1.37\\
    \bottomrule
  \end{tabular}
  }
   \vspace{-0.4cm}
\end{table}

\subsection{Further Analyses}
To answer \textbf{Q2}, in this section, we conduct further analyses to verify the designs of our reward functions, phase-dependent training strategy, and hybrid retrieval. We also discuss the general compatibility of our method with various retrieval methods to answer \textbf{Q3}. 
\subsubsection{PRA \& CAF Rewards}
We perform ablation studies to evaluate the contribution of PRA and CAF reward designs. We remove PRA, CAF, and both rewards, denoted as ``-PRA'', ``-CAF'', and ``-PRA \& CAF'', respectively, and present the results in Table~\ref{tab:pra-caf}. The results show that removing either PRA or CAF, or both, leads to substantial performance drops in terms of F1 score. This confirms that both mechanisms are indispensable components of \methodname~. 

\begin{table}
\setlength{\tabcolsep}{2.5pt}
  \caption{The results of the ablation study on the phase-dependent training strategy.}
  \vspace{-0.4cm}
  \label{tab:cold}
  \resizebox{0.48\textwidth}{!}{
  \begin{tabular}{lccccccccc}
    \toprule
    \multirow{2}{*}{Method}     & \multicolumn{2}{c}{HotpotQA}    & \multicolumn{2}{c}{MuSiQue}    & \multicolumn{2}{c}{2Wiki} & \multicolumn{2}{c}{PopQA}\\

              \cmidrule[1pt](lr){2-3}\cmidrule[1pt](lr){4-5}\cmidrule[1pt](lr){6-7}\cmidrule[1pt](lr){8-9}\
                             & $\mathrm{F1}$                          & $\mathrm{ACC_{F}}$                             & $\mathrm{F1}$                           & $\mathrm{ACC_{F}}$                          & $\mathrm{F1}$                          & $\mathrm{ACC_{F}}$               & $\mathrm{F1}$                          & $\mathrm{ACC_{F}}$       \\
    \midrule
        \text{\methodname} &    \textbf{38.00}     &     \textbf{92.00}     &   \textbf{20.06}   &  \textbf{89.50}       &    \textbf{32.24}    &   \textbf{89.00}  & \textbf{35.04} &  \textbf{95.10} \\
        \text{- cold start}       &    \underline{30.31}   &     \underline{89.00}     &   14.84   &  \underline{89.50}       &    \underline{30.04}   &   \underline{89.50}  & 28.83&   \underline{92.80} \\

        \text{- all stages} &    26.12  &     71.00     &   \underline{16.79}    &  74.00      &    28.05   &   76.50  & \underline{30.23} &  82.50 \\

    \bottomrule
  \end{tabular}
  }
  \vspace{-0.4cm}
\end{table}

Besides the quantitative F1 score, we also delve deeper into the results by reporting the number of retrieval calls. We make the following observations. Without using PRA, our method has a reduced number of retrieval calls, indicating that it suffers from the ``shallow retrieval'' problem and leads to unsatisfactory performance. PRA makes the retrieval behavior more controllable. 

\begin{table*}
\caption{The results of integrating \methodname~with other GraphRAG retrieval methods. Note that \methodname~is only trained once and do not access other RAG methods during training. Imp denotes the average improvement across all datasets. }
\vspace{-0.4cm}
  \label{tab:generalization}
  \setlength{\tabcolsep}{3.2pt}
  \begin{tabular}{clccccccccccccc}
    \toprule
    \multirow{2}{*}{Method} & \multirow{2}{*}{Model}      & \multicolumn{3}{c} {HotpotQA}    & \multicolumn{3}{c}{MuSiQue}    & \multicolumn{3}{c}{2Wiki}  & \multicolumn{3}{c}{PopQA}   & \multirow{2}{*}{Imp}\\

              \cmidrule[1pt](lr){3-5}\cmidrule[1pt](lr){6-8}\cmidrule[1pt](lr){9-11}\cmidrule[1pt](lr){12-14}\
                         &    & $\mathrm{F1}$                          & $\mathrm{ACC_{L}}$               & $\mathrm{SBERT}$       & $\mathrm{F1}$                           & $\mathrm{ACC_{L}}$                 & $\mathrm{SBERT}$     & $\mathrm{F1}$                          & $\mathrm{ACC_{L}}$                   & $\mathrm{SBERT}$   & $\mathrm{F1}$  & $\mathrm{ACC_{L}}$                  & $\mathrm{SBERT}$  \\
  \midrule
       \multirow{2}{*}{Naive RAG}  &   Original & 24.24  &   \textbf{49.00}   &  58.73 &  8.99   & 19.00   &  48.97  &   13.14  &  25.00   &  52.56  &   23.72    &   \textbf{58.10}   &     58.60  &  \\
    &   +\methodname &  \textbf{32.08} &  48.00   &   \textbf{62.72}   &  \textbf{20.77} &    \textbf{28.00}   & \textbf{56.59}   & \textbf{30.14} &    \textbf{37.50}   & \textbf{61.22}   & \textbf{31.96}     & 57.50&\textbf{52.96}& + 18.04\% \\

    \midrule
    
    \multirow{2}{*}{KGP}&   Original &  10.73   &   21.00    &  50.92  &   4.61  &   11.00  &  46.70   & 10.16   & 18.00    &  50.27   & \textbf{21.01} & 50.00 &\textbf{56.10} \\
    &   +\methodname &  \textbf{36.16}  &  \textbf{41.50}  &  \textbf{65.81}  &  \textbf{13.33}  &      \textbf{ 14.50}  &  \textbf{53.82} &  \textbf{27.79} &   \textbf{35.50}  &  \textbf{60.41} &  18.33  &  \textbf{55.00}  & 55.39 & + 36.25\% \\

   \midrule
       
    \multirow{2}{*}{ToG}&   Original &  11.44 &  21.50   &   50.48 &  5.02 &   8.00     &  47.05 &  14.45 &     21.50   &  53.11  &  \textbf{29.21} &  \textbf{47.00}  &  \textbf{62.94}   &   \\
    &   +\methodname & \textbf{16.82}   &\textbf{25.50}   &   \textbf{55.36}   &   \textbf{8.29} &    \textbf{8.50}    &   \textbf{51.77}  &  \textbf{21.95} &  \textbf{26.00}    &  \textbf{57.68}   & 22.30 &   37.00 & 60.10 & + 5.26\%  \\

    \midrule
    
    \multirow{2}{*}{LightRAG}&   Original &   7.96  & 18.50  &   49.15 &   3.09 & 6.00   &  44.99  &   7.58  &  10.00   &  47.95 &  14.55&  34.30  &   53.07 &  \\
    &   +\methodname & \textbf{20.85}  &  \textbf{27.00} &  \textbf{57.40}  & \textbf{8.59}  &   \textbf{12.50}    &  \textbf{48.95}  & \textbf{19.64}  &    \textbf{23.00}  &  \textbf{55.31}   & \textbf{28.42}     & \textbf{49.00} &\textbf{60.50}& + 38.37\% \\

       \midrule

    \multirow{2}{*}{PropRAG}&   Original   &  25.47 &  \textbf{55.00}  &   59.76  &  10.11 &    \textbf{21.50}  &  50.77 &   17.17 &   35.00  &  53.85 &23.29  &  57.30  & 58.22  &  \\
    &  +\methodname &     \textbf{30.87} & 42.00  &   \textbf{62.25}  & \textbf{18.78} &    21.00  & \textbf{55.32}  &  \textbf{34.86} &   \textbf{45.50}  &   \textbf{64.44} &  \textbf{32.69}    &\textbf{57.80} &\textbf{63.44}&  + 13.16\% \\
 
       \midrule
       
    \multirow{2}{*}{G-retriever}&   Original &   \textbf{14.81}  &  \textbf{21.50}  &   \textbf{54.89}  &   3.38 &   \textbf{5.00}  &  47.47 &  15.12   &   \textbf{19.00} &   52.77  &  2.14  &    6.80  &   44.63 &    \\
    &   +\methodname  &  13.52  &  15.50 &    54.64     &  \textbf{10.56} &   3.50   &  \textbf{54.73}  &   \textbf{15.23} &  17.50   &  \textbf{54.07}  &  \textbf{5.86}  &\textbf{11.10} &\textbf{49.06}& + 6.18\%  \\

       \midrule
       
    \multirow{2}{*}{HippoRAG2}&   Original&   27.52  &  \textbf{53.00}  & 61.02    &  8.90  &  23.50   & 50.19  & 12.32     & 29.00   & 51.92   &  23.99&  57.46 &   58.16     \\

    &   +\methodname  &  \textbf{38.00} &   52.00   & \textbf{66.66}   & \textbf{20.06}  &  \textbf{26.50}    & \textbf{57.49} &  \textbf{32.24}  &  \textbf{42.50}   & \textbf{63.56}       &   \textbf{35.04}    &    \textbf{60.00}  & \textbf{65.31}& + 22.40\% \\

    \bottomrule
  \end{tabular}
  \vspace{-0.4cm}
\end{table*}

On the other hand, without using CAF, the number of retrieval calls consistently increases across all datasets, indicating the model is more inclined to invoke external functions. However, this not yield to corresponding F1 improvements in answer quality. In fact, the F1 score is even lower than methods invoking few retrievals, indicating that the model tends towards ``over-thinking'' and performing unnecessary retrievals.

When both rewards are adopted, our method achieves the best performance with a suitable number of retrieval calls, indicating our method can calibrate its retrieval strategy to the inherent difficulty of the problem. For HotpotQA and PopQA, which contain relatively simpler problems, the number of retrieval calls is automatically decreased. Conversely, for inherently more complex problems demanding deeper information gathering, such as MuSiQue and 2Wiki, our model learns to invoke retrievals more frequently.

\subsubsection{Phase-Dependent Training Strategy}
To validate the effectiveness of our training strategy, we compare our method with two variants: one without the first cold-start stage (denoted as ``-cold start''), and the other without distinct stages, i.e., directly mixing all reward functions for training (denoted as ``-all stages''). The results are shown in Table ~\ref{tab:cold}. Here, $\mathrm{ACC_{F}}$ represents the percentage of model outputs that are entirely accurate in format. %

Introducing the cold-start strategy effectively improves performance. %
In this stage, the model can learn simple tasks like format learning and retrieval invocation, which is beneficial for optimizing advanced policies in the later two stages. 

Compared to training, which used format, PRA, and CAF rewards simultaneously, we found that the three-stage training strategy delivers superior performance. 
This approach avoids conflicts or convergence difficulties caused by overly complex reward signals.

\subsubsection{Hybrid Graph-Textual Retrieval}\label{sec:exp:hybrid}
\begin{figure}[t]
	\centering
		\includegraphics[width=0.49\columnwidth]{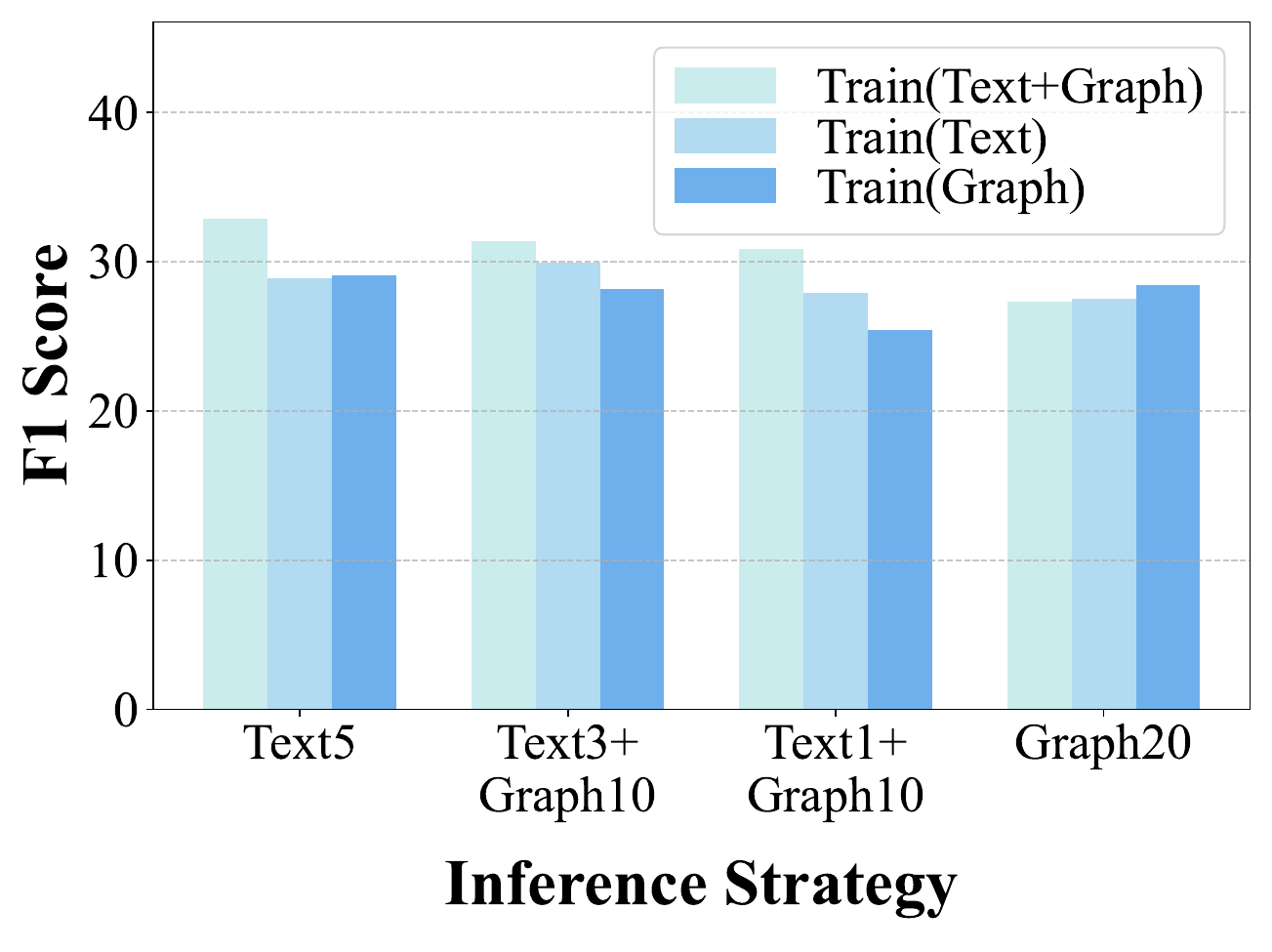}
		\includegraphics[width=0.49\columnwidth]{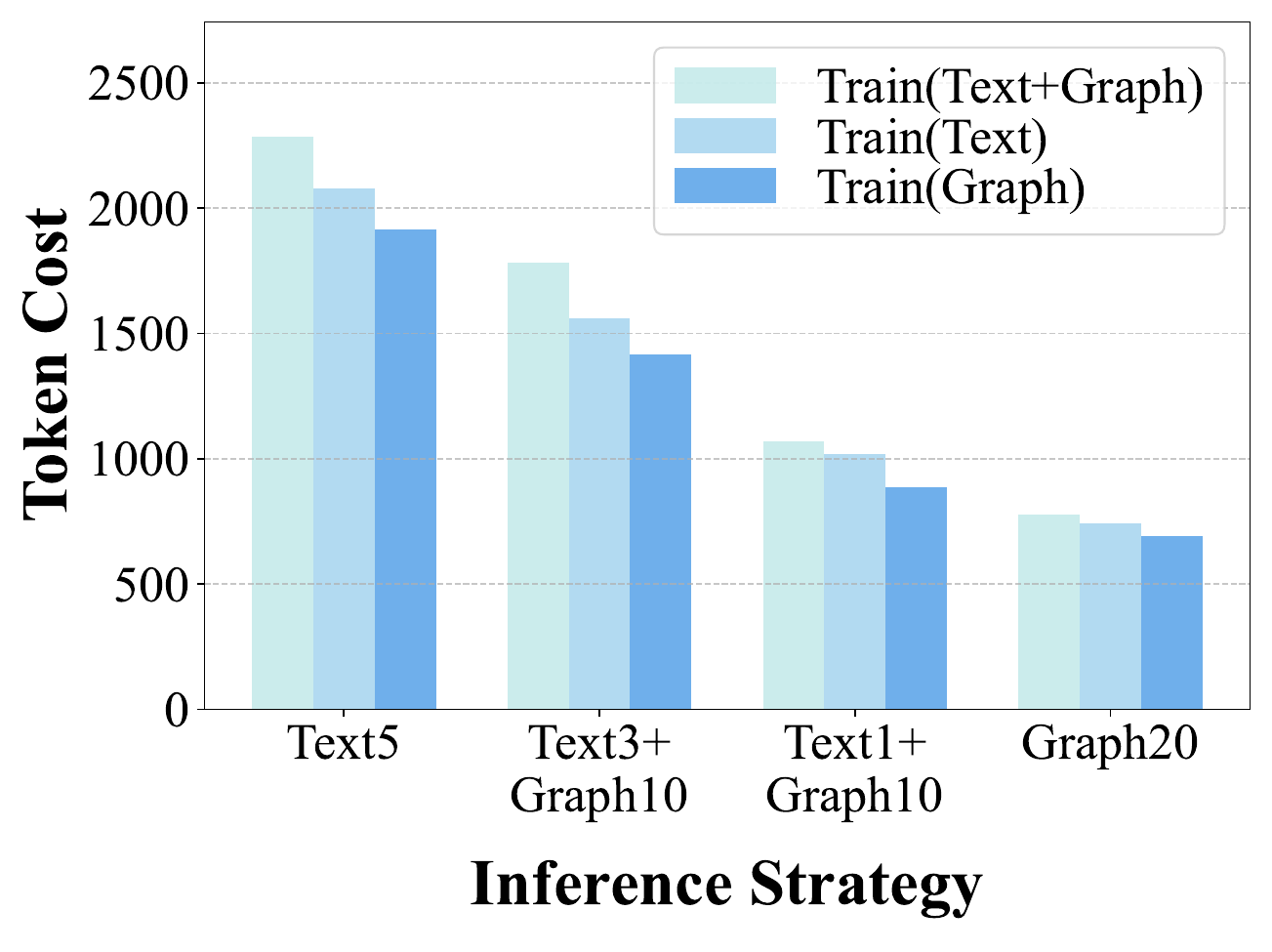}
        \vspace{-0.5cm}
        \caption{A comparison of different retrieval formats. Left: the F1 score comparison. Right: the token cost for testing.}\label{fig:tuple3}
        \Description{This figure presents a comparative analysis of different knowledge retrieval strategies within the GraphRAG-R1 framework, evaluating their effectiveness and computational efficiency through two key metrics displayed in side-by-side sub-figures:

Left Sub-figure (F1 Score Comparison): This bar chart compares the answer accuracy, measured by the F1 score, achieved when using different retrieval formats or mixtures during inference. It typically evaluates strategies such as using only text snippets, only graph triplets, or various hybrid combinations (e.g., Text5, Text3+Graph10, Text1+Graph10, Graph20). The results visually demonstrate how the hybrid retrieval design balances and optimizes performance.

Right Sub-figure (Token Cost for Testing): This accompanying bar chart (or line chart) quantifies the computational cost associated with each retrieval strategy shown on the left. It measures the average number of tokens consumed during the testing phase, directly relating to inference time and resource usage. This sub-figure highlights the efficiency gains of using structured graph information, which is typically more information-dense than pure text.}
        \vspace{-0.6cm}
\end{figure}

To demonstrate the effectiveness of our hybrid graph-textual retrieval design, we compare different retrieval strategies. During training, we adopt three settings: using mixing text and graph (``Text+Graph''), text retrieval only (``Text''), and using graph only ("Graph"). During inference, we utilize 4 different retrieval methods: using 5 text snippets (``Text5''), using 3 text snippets and 10 graph triplets (``Text3+Graph10''), using 1 text snippet and 10 graph triplets (``Text1+Graph10''), and using 20 graph triplets ( ``Graph20''). 
The results are shown in Figure ~\ref{fig:tuple3}.

For different training strategies, training only with texts and training only with graph generally lead to performance decrease in the F1 score, demonstrating the effectiveness of mixed training.

For different inference strategies, as the proportion of graph data increases, the number of tokens consumed significantly reduces, while the F1 score of the model only slightly decreases. 
This demonstrates that our hybrid retrieval approach can achieve a suitable balance between model effectiveness and computational resource consumption.  %

\subsubsection{Integration with Other Retrievers}
To answer \textbf{Q3}, we conduct experiments to integrate our proposed method with other existing retrieval methods. 
As shown in Table \ref{tab:generalization}, \methodname~generally improves the performance of different RAG methods across datasets. The last column "Imp" is defined as the average percentage increase in all metrics after integrating \methodname~ relative to original method. 
Specifically, GraphRAG-R1 achieves notable average improvements over the original methods, with the highest improvement of 38.37\% on LightRAG and consistent gains ranging from 5.26\% to 36.25\% in other baselines. Note that our method is only trained once and only has access to HippoRAG2 during training, indicating that the reasoning and retrieval ability learned by \methodname~can be easily generalized to other GraphRAG methods.
More experiments are provided in the Appendix.

\section{Conclusion}
In this paper, we propose \methodname, an adaptive GraphRAG framework to enhance the multi-hop reasoning ability of LLMs through process-constrained RL. 
Our core designs include the \rlname~training method, \rewardname~designs including PRA, CAF, and phase-depending training, and a hybrid graph-textual retrieval. 
Experimental results validate the superiority of \methodname~over state-of-the-art baselines across both in-domain and out-of-domain datasets. We also demonstrate our method can be flexibly integrated with other retrieval methods.  %
Future works include applying our method to more types of graphs and application scenarios.

\begin{acks}
This work was supported in part by the National Natural Science Foundation of China (No. 92570113, 62472018), the Natural Science Foundation of Tianjin (No. 24JCQNJC02170), and the Tianjin Science and Technology Plan Project (No. 24HHXCSS00001).
\end{acks}

\bibliographystyle{ACM-Reference-Format}
\bibliography{sample-base}


\begin{thebibliography}{54}


\ifx \showCODEN    \undefined \def \showCODEN     #1{\unskip}     \fi
\ifx \showISBNx    \undefined \def \showISBNx     #1{\unskip}     \fi
\ifx \showISBNxiii \undefined \def \showISBNxiii  #1{\unskip}     \fi
\ifx \showISSN     \undefined \def \showISSN      #1{\unskip}     \fi
\ifx \showLCCN     \undefined \def \showLCCN      #1{\unskip}     \fi
\ifx \shownote     \undefined \def \shownote      #1{#1}          \fi
\ifx \showarticletitle \undefined \def \showarticletitle #1{#1}   \fi
\ifx \showURL      \undefined \def \showURL       {\relax}        \fi
\providecommand\bibfield[2]{#2}
\providecommand\bibinfo[2]{#2}
\providecommand\natexlab[1]{#1}
\providecommand\showeprint[2][]{arXiv:#2}

\bibitem[Asai et~al\mbox{.}(2024)]%
        {asai2024self}
\bibfield{author}{\bibinfo{person}{Akari Asai}, \bibinfo{person}{Zeqiu Wu}, \bibinfo{person}{Yizhong Wang}, \bibinfo{person}{Avirup Sil}, {and} \bibinfo{person}{Hannaneh Hajishirzi}.} \bibinfo{year}{2024}\natexlab{}.
\newblock \showarticletitle{Self-rag: Learning to retrieve, generate, and critique through self-reflection}.
\newblock  (\bibinfo{year}{2024}).
\newblock


\bibitem[Chen et~al\mbox{.}(2024)]%
        {chen2024not}
\bibfield{author}{\bibinfo{person}{Xingyu Chen}, \bibinfo{person}{Jiahao Xu}, \bibinfo{person}{Tian Liang}, \bibinfo{person}{Zhiwei He}, \bibinfo{person}{Jianhui Pang}, \bibinfo{person}{Dian Yu}, \bibinfo{person}{Linfeng Song}, \bibinfo{person}{Qiuzhi Liu}, \bibinfo{person}{Mengfei Zhou}, \bibinfo{person}{Zhuosheng Zhang}, {et~al\mbox{.}}} \bibinfo{year}{2024}\natexlab{}.
\newblock \showarticletitle{Do not think that much for 2+ 3=? on the overthinking of o1-like llms}.
\newblock \bibinfo{journal}{\emph{arXiv preprint arXiv:2412.21187}} (\bibinfo{year}{2024}).
\newblock


\bibitem[Cuesta-Ramirez et~al\mbox{.}(2025)]%
        {cuesta2025large}
\bibfield{author}{\bibinfo{person}{Jhouben Cuesta-Ramirez}, \bibinfo{person}{Samuel Beaussant}, {and} \bibinfo{person}{Mehdi Mounsif}.} \bibinfo{year}{2025}\natexlab{}.
\newblock \showarticletitle{Large Reasoning Models are not thinking straight: on the unreliability of thinking trajectories}.
\newblock \bibinfo{journal}{\emph{arXiv preprint arXiv:2507.00711}} (\bibinfo{year}{2025}).
\newblock


\bibitem[Deng et~al\mbox{.}(2025)]%
        {deng2025boosting}
\bibfield{author}{\bibinfo{person}{Huilin Deng}, \bibinfo{person}{Ding Zou}, \bibinfo{person}{Rui Ma}, \bibinfo{person}{Hongchen Luo}, \bibinfo{person}{Yang Cao}, {and} \bibinfo{person}{Yu Kang}.} \bibinfo{year}{2025}\natexlab{}.
\newblock \showarticletitle{Boosting the generalization and reasoning of vision language models with curriculum reinforcement learning}.
\newblock \bibinfo{journal}{\emph{arXiv preprint arXiv:2503.07065}} (\bibinfo{year}{2025}).
\newblock


\bibitem[Edge et~al\mbox{.}(2024)]%
        {micro_graphrag}
\bibfield{author}{\bibinfo{person}{Darren Edge}, \bibinfo{person}{Ha Trinh}, \bibinfo{person}{Newman Cheng}, \bibinfo{person}{Joshua Bradley}, \bibinfo{person}{Alex Chao}, \bibinfo{person}{Apurva Mody}, \bibinfo{person}{Steven Truitt}, \bibinfo{person}{Dasha Metropolitansky}, \bibinfo{person}{Robert~Osazuwa Ness}, {and} \bibinfo{person}{Jonathan Larson}.} \bibinfo{year}{2024}\natexlab{}.
\newblock \showarticletitle{From local to global: A graph rag approach to query-focused summarization}.
\newblock \bibinfo{journal}{\emph{arXiv preprint arXiv:2404.16130}} (\bibinfo{year}{2024}).
\newblock


\bibitem[Fan et~al\mbox{.}(2024)]%
        {fan2024survey}
\bibfield{author}{\bibinfo{person}{Wenqi Fan}, \bibinfo{person}{Yujuan Ding}, \bibinfo{person}{Liangbo Ning}, \bibinfo{person}{Shijie Wang}, \bibinfo{person}{Hengyun Li}, \bibinfo{person}{Dawei Yin}, \bibinfo{person}{Tat-Seng Chua}, {and} \bibinfo{person}{Qing Li}.} \bibinfo{year}{2024}\natexlab{}.
\newblock \showarticletitle{A survey on rag meeting llms: Towards retrieval-augmented large language models}. In \bibinfo{booktitle}{\emph{Proceedings of the 30th ACM SIGKDD conference on knowledge discovery and data mining}}. \bibinfo{pages}{6491--6501}.
\newblock


\bibitem[Gao et~al\mbox{.}(2023a)]%
        {gao2023scaling}
\bibfield{author}{\bibinfo{person}{Leo Gao}, \bibinfo{person}{John Schulman}, {and} \bibinfo{person}{Jacob Hilton}.} \bibinfo{year}{2023}\natexlab{a}.
\newblock \showarticletitle{Scaling laws for reward model overoptimization}. In \bibinfo{booktitle}{\emph{International Conference on Machine Learning}}. PMLR, \bibinfo{pages}{10835--10866}.
\newblock


\bibitem[Gao et~al\mbox{.}(2023b)]%
        {gao2023retrieval}
\bibfield{author}{\bibinfo{person}{Yunfan Gao}, \bibinfo{person}{Yun Xiong}, \bibinfo{person}{Xinyu Gao}, \bibinfo{person}{Kangxiang Jia}, \bibinfo{person}{Jinliu Pan}, \bibinfo{person}{Yuxi Bi}, \bibinfo{person}{Yixin Dai}, \bibinfo{person}{Jiawei Sun}, \bibinfo{person}{Haofen Wang}, {and} \bibinfo{person}{Haofen Wang}.} \bibinfo{year}{2023}\natexlab{b}.
\newblock \showarticletitle{Retrieval-augmented generation for large language models: A survey}.
\newblock \bibinfo{journal}{\emph{arXiv preprint arXiv:2312.10997}} \bibinfo{volume}{2}, \bibinfo{number}{1} (\bibinfo{year}{2023}).
\newblock


\bibitem[Guan et~al\mbox{.}(2025)]%
        {deeprag}
\bibfield{author}{\bibinfo{person}{Xinyan Guan}, \bibinfo{person}{Jiali Zeng}, \bibinfo{person}{Fandong Meng}, \bibinfo{person}{Chunlei Xin}, \bibinfo{person}{Yaojie Lu}, \bibinfo{person}{Hongyu Lin}, \bibinfo{person}{Xianpei Han}, \bibinfo{person}{Le Sun}, {and} \bibinfo{person}{Jie Zhou}.} \bibinfo{year}{2025}\natexlab{}.
\newblock \showarticletitle{DeepRAG: Thinking to Retrieve Step by Step for Large Language Models}.
\newblock \bibinfo{journal}{\emph{arXiv preprint arXiv:2502.01142}} (\bibinfo{year}{2025}).
\newblock


\bibitem[Guo et~al\mbox{.}(2025)]%
        {guo2025deepseek}
\bibfield{author}{\bibinfo{person}{Daya Guo}, \bibinfo{person}{Dejian Yang}, \bibinfo{person}{Haowei Zhang}, \bibinfo{person}{Junxiao Song}, \bibinfo{person}{Ruoyu Zhang}, \bibinfo{person}{Runxin Xu}, \bibinfo{person}{Qihao Zhu}, \bibinfo{person}{Shirong Ma}, \bibinfo{person}{Peiyi Wang}, \bibinfo{person}{Xiao Bi}, {et~al\mbox{.}}} \bibinfo{year}{2025}\natexlab{}.
\newblock \showarticletitle{Deepseek-r1: Incentivizing reasoning capability in llms via reinforcement learning}.
\newblock \bibinfo{journal}{\emph{arXiv preprint arXiv:2501.12948}} (\bibinfo{year}{2025}).
\newblock


\bibitem[Guo et~al\mbox{.}(2024)]%
        {guo2024lightrag}
\bibfield{author}{\bibinfo{person}{Zirui Guo}, \bibinfo{person}{Lianghao Xia}, \bibinfo{person}{Yanhua Yu}, \bibinfo{person}{Tu Ao}, {and} \bibinfo{person}{Chao Huang}.} \bibinfo{year}{2024}\natexlab{}.
\newblock \showarticletitle{Lightrag: Simple and fast retrieval-augmented generation}.
\newblock \bibinfo{journal}{\emph{arXiv preprint arXiv:2410.05779}} (\bibinfo{year}{2024}).
\newblock


\bibitem[Guti{\'e}rrez et~al\mbox{.}(2025)]%
        {gutierrez2025hipporag2}
\bibfield{author}{\bibinfo{person}{Bernal~Jim{\'e}nez Guti{\'e}rrez}, \bibinfo{person}{Yiheng Shu}, \bibinfo{person}{Weijian Qi}, \bibinfo{person}{Sizhe Zhou}, {and} \bibinfo{person}{Yu Su}.} \bibinfo{year}{2025}\natexlab{}.
\newblock \showarticletitle{From rag to memory: Non-parametric continual learning for large language models}.
\newblock \bibinfo{journal}{\emph{arXiv preprint arXiv:2502.14802}} (\bibinfo{year}{2025}).
\newblock


\bibitem[Han et~al\mbox{.}(2024)]%
        {hansurvey}
\bibfield{author}{\bibinfo{person}{Haoyu Han}, \bibinfo{person}{Yu Wang}, \bibinfo{person}{Harry Shomer}, \bibinfo{person}{Kai Guo}, \bibinfo{person}{Jiayuan Ding}, \bibinfo{person}{Yongjia Lei}, \bibinfo{person}{Mahantesh Halappanavar}, \bibinfo{person}{Ryan~A Rossi}, \bibinfo{person}{Subhabrata Mukherjee}, \bibinfo{person}{Xianfeng Tang}, {et~al\mbox{.}}} \bibinfo{year}{2024}\natexlab{}.
\newblock \showarticletitle{Retrieval-augmented generation with graphs (graphrag)}.
\newblock \bibinfo{journal}{\emph{arXiv preprint arXiv:2501.00309}} (\bibinfo{year}{2024}).
\newblock


\bibitem[He et~al\mbox{.}(2024)]%
        {he2024grtriever}
\bibfield{author}{\bibinfo{person}{Xiaoxin He}, \bibinfo{person}{Yijun Tian}, \bibinfo{person}{Yifei Sun}, \bibinfo{person}{Nitesh Chawla}, \bibinfo{person}{Thomas Laurent}, \bibinfo{person}{Yann LeCun}, \bibinfo{person}{Xavier Bresson}, {and} \bibinfo{person}{Bryan Hooi}.} \bibinfo{year}{2024}\natexlab{}.
\newblock \showarticletitle{G-retriever: Retrieval-augmented generation for textual graph understanding and question answering}.
\newblock \bibinfo{journal}{\emph{Advances in Neural Information Processing Systems}}  \bibinfo{volume}{37} (\bibinfo{year}{2024}), \bibinfo{pages}{132876--132907}.
\newblock


\bibitem[Helwe et~al\mbox{.}(2021)]%
        {helwe2021reasoning}
\bibfield{author}{\bibinfo{person}{Chadi Helwe}, \bibinfo{person}{Chlo{\'e} Clavel}, {and} \bibinfo{person}{Fabian Suchanek}.} \bibinfo{year}{2021}\natexlab{}.
\newblock \showarticletitle{Reasoning with transformer-based models: Deep learning, but shallow reasoning}. In \bibinfo{booktitle}{\emph{2021 International Conference on Automated Knowledge Base Construction (AKBC)}}.
\newblock


\bibitem[Ho et~al\mbox{.}(2020)]%
        {ho2020constructing}
\bibfield{author}{\bibinfo{person}{Xanh Ho}, \bibinfo{person}{Anh-Khoa~Duong Nguyen}, \bibinfo{person}{Saku Sugawara}, {and} \bibinfo{person}{Akiko Aizawa}.} \bibinfo{year}{2020}\natexlab{}.
\newblock \showarticletitle{Constructing A Multi-hop QA Dataset for Comprehensive Evaluation of Reasoning Steps}. In \bibinfo{booktitle}{\emph{Proceedings of the 28th International Conference on Computational Linguistics}}. \bibinfo{pages}{6609--6625}.
\newblock


\bibitem[Hu et~al\mbox{.}(2021)]%
        {hu2021lora}
\bibfield{author}{\bibinfo{person}{Edward Hu}, \bibinfo{person}{Yelong Shen}, \bibinfo{person}{Phillip Wallis}, \bibinfo{person}{Zeyuan Allen-Zhu}, \bibinfo{person}{Yuanzhi Li}, \bibinfo{person}{Shean Wang}, {and} \bibinfo{person}{Weizhu Chen}.} \bibinfo{year}{2021}\natexlab{}.
\newblock \showarticletitle{LoRA: Low-Rank Adaptation of Large Language Models}.
\newblock  (\bibinfo{year}{2021}).
\newblock


\bibitem[Jaech et~al\mbox{.}(2024)]%
        {jaech2024openai}
\bibfield{author}{\bibinfo{person}{Aaron Jaech}, \bibinfo{person}{Adam Kalai}, \bibinfo{person}{Adam Lerer}, \bibinfo{person}{Adam Richardson}, \bibinfo{person}{Ahmed El-Kishky}, \bibinfo{person}{Aiden Low}, \bibinfo{person}{Alec Helyar}, \bibinfo{person}{Aleksander Madry}, \bibinfo{person}{Alex Beutel}, \bibinfo{person}{Alex Carney}, {et~al\mbox{.}}} \bibinfo{year}{2024}\natexlab{}.
\newblock \showarticletitle{Openai o1 system card}.
\newblock \bibinfo{journal}{\emph{arXiv preprint arXiv:2412.16720}} (\bibinfo{year}{2024}).
\newblock


\bibitem[Jiang and Ferrara(2023)]%
        {jiang2023social}
\bibfield{author}{\bibinfo{person}{Julie Jiang} {and} \bibinfo{person}{Emilio Ferrara}.} \bibinfo{year}{2023}\natexlab{}.
\newblock \showarticletitle{Social-llm: Modeling user behavior at scale using language models and social network data}.
\newblock \bibinfo{journal}{\emph{arXiv preprint arXiv:2401.00893}} (\bibinfo{year}{2023}).
\newblock


\bibitem[Jimenez~Gutierrez et~al\mbox{.}(2024)]%
        {jimenez2024hipporag}
\bibfield{author}{\bibinfo{person}{Bernal Jimenez~Gutierrez}, \bibinfo{person}{Yiheng Shu}, \bibinfo{person}{Yu Gu}, \bibinfo{person}{Michihiro Yasunaga}, {and} \bibinfo{person}{Yu Su}.} \bibinfo{year}{2024}\natexlab{}.
\newblock \showarticletitle{Hipporag: Neurobiologically inspired long-term memory for large language models}.
\newblock \bibinfo{journal}{\emph{Advances in Neural Information Processing Systems}}  \bibinfo{volume}{37} (\bibinfo{year}{2024}), \bibinfo{pages}{59532--59569}.
\newblock


\bibitem[Jin et~al\mbox{.}(2025)]%
        {jin2025searchr1}
\bibfield{author}{\bibinfo{person}{Bowen Jin}, \bibinfo{person}{Hansi Zeng}, \bibinfo{person}{Zhenrui Yue}, \bibinfo{person}{Jinsung Yoon}, \bibinfo{person}{Sercan Arik}, \bibinfo{person}{Dong Wang}, \bibinfo{person}{Hamed Zamani}, {and} \bibinfo{person}{Jiawei Han}.} \bibinfo{year}{2025}\natexlab{}.
\newblock \showarticletitle{Search-r1: Training llms to reason and leverage search engines with reinforcement learning}.
\newblock \bibinfo{journal}{\emph{arXiv preprint arXiv:2503.09516}} (\bibinfo{year}{2025}).
\newblock


\bibitem[Li et~al\mbox{.}(2024b)]%
        {li2024dalk}
\bibfield{author}{\bibinfo{person}{Dawei Li}, \bibinfo{person}{Shu Yang}, \bibinfo{person}{Zhen Tan}, \bibinfo{person}{Jae Baik}, \bibinfo{person}{Sukwon Yun}, \bibinfo{person}{Joseph Lee}, \bibinfo{person}{Aaron Chacko}, \bibinfo{person}{Bojian Hou}, \bibinfo{person}{Duy Duong-Tran}, \bibinfo{person}{Ying Ding}, {et~al\mbox{.}}} \bibinfo{year}{2024}\natexlab{b}.
\newblock \showarticletitle{DALK: Dynamic Co-Augmentation of LLMs and KG to answer Alzheimer’s Disease Questions with Scientific Literature}. In \bibinfo{booktitle}{\emph{Findings of the Association for Computational Linguistics: EMNLP 2024}}. \bibinfo{pages}{2187--2205}.
\newblock


\bibitem[Li et~al\mbox{.}(2024a)]%
        {li2024graphreader}
\bibfield{author}{\bibinfo{person}{Shilong Li}, \bibinfo{person}{Yancheng He}, \bibinfo{person}{Hangyu Guo}, \bibinfo{person}{Xingyuan Bu}, \bibinfo{person}{Ge Bai}, \bibinfo{person}{Jie Liu}, \bibinfo{person}{Jiaheng Liu}, \bibinfo{person}{Xingwei Qu}, \bibinfo{person}{Yangguang Li}, \bibinfo{person}{Wanli Ouyang}, {et~al\mbox{.}}} \bibinfo{year}{2024}\natexlab{a}.
\newblock \showarticletitle{GraphReader: Building Graph-based Agent to Enhance Long-Context Abilities of Large Language Models}. In \bibinfo{booktitle}{\emph{Findings of the Association for Computational Linguistics: EMNLP 2024}}. \bibinfo{pages}{12758--12786}.
\newblock


\bibitem[Li et~al\mbox{.}(2025)]%
        {li2025g}
\bibfield{author}{\bibinfo{person}{Yuhan Li}, \bibinfo{person}{Xinni Zhang}, \bibinfo{person}{Linhao Luo}, \bibinfo{person}{Heng Chang}, \bibinfo{person}{Yuxiang Ren}, \bibinfo{person}{Irwin King}, {and} \bibinfo{person}{Jia Li}.} \bibinfo{year}{2025}\natexlab{}.
\newblock \showarticletitle{G-refer: Graph retrieval-augmented large language model for explainable recommendation}. In \bibinfo{booktitle}{\emph{Proceedings of the ACM on Web Conference 2025}}. \bibinfo{pages}{240--251}.
\newblock


\bibitem[Ma et~al\mbox{.}(2024)]%
        {tog2}
\bibfield{author}{\bibinfo{person}{Shengjie Ma}, \bibinfo{person}{Chengjin Xu}, \bibinfo{person}{Xuhui Jiang}, \bibinfo{person}{Muzhi Li}, \bibinfo{person}{Huaren Qu}, \bibinfo{person}{Cehao Yang}, \bibinfo{person}{Jiaxin Mao}, {and} \bibinfo{person}{Jian Guo}.} \bibinfo{year}{2024}\natexlab{}.
\newblock \showarticletitle{Think-on-graph 2.0: Deep and faithful large language model reasoning with knowledge-guided retrieval augmented generation}.
\newblock \bibinfo{journal}{\emph{arXiv preprint arXiv:2407.10805}} (\bibinfo{year}{2024}).
\newblock


\bibitem[Mallen et~al\mbox{.}(2023)]%
        {mallen2023not}
\bibfield{author}{\bibinfo{person}{Alex Mallen}, \bibinfo{person}{Akari Asai}, \bibinfo{person}{Victor Zhong}, \bibinfo{person}{Rajarshi Das}, \bibinfo{person}{Daniel Khashabi}, {and} \bibinfo{person}{Hannaneh Hajishirzi}.} \bibinfo{year}{2023}\natexlab{}.
\newblock \showarticletitle{When Not to Trust Language Models: Investigating Effectiveness of Parametric and Non-Parametric Memories}. In \bibinfo{booktitle}{\emph{Proceedings of the 61st Annual Meeting of the Association for Computational Linguistics (Volume 1: Long Papers)}}. \bibinfo{pages}{9802--9822}.
\newblock


\bibitem[Ouyang et~al\mbox{.}(2022)]%
        {ouyang2022training}
\bibfield{author}{\bibinfo{person}{Long Ouyang}, \bibinfo{person}{Jeffrey Wu}, \bibinfo{person}{Xu Jiang}, \bibinfo{person}{Diogo Almeida}, \bibinfo{person}{Carroll Wainwright}, \bibinfo{person}{Pamela Mishkin}, \bibinfo{person}{Chong Zhang}, \bibinfo{person}{Sandhini Agarwal}, \bibinfo{person}{Katarina Slama}, \bibinfo{person}{Alex Ray}, {et~al\mbox{.}}} \bibinfo{year}{2022}\natexlab{}.
\newblock \showarticletitle{Training language models to follow instructions with human feedback}.
\newblock \bibinfo{journal}{\emph{Advances in neural information processing systems}}  \bibinfo{volume}{35} (\bibinfo{year}{2022}), \bibinfo{pages}{27730--27744}.
\newblock


\bibitem[Pan et~al\mbox{.}(2024)]%
        {pan2024unifying}
\bibfield{author}{\bibinfo{person}{Shirui Pan}, \bibinfo{person}{Linhao Luo}, \bibinfo{person}{Yufei Wang}, \bibinfo{person}{Chen Chen}, \bibinfo{person}{Jiapu Wang}, {and} \bibinfo{person}{Xindong Wu}.} \bibinfo{year}{2024}\natexlab{}.
\newblock \showarticletitle{Unifying large language models and knowledge graphs: A roadmap}.
\newblock \bibinfo{journal}{\emph{IEEE Transactions on Knowledge and Data Engineering}} \bibinfo{volume}{36}, \bibinfo{number}{7} (\bibinfo{year}{2024}), \bibinfo{pages}{3580--3599}.
\newblock


\bibitem[Peng et~al\mbox{.}(2024)]%
        {pengsurvey}
\bibfield{author}{\bibinfo{person}{Boci Peng}, \bibinfo{person}{Yun Zhu}, \bibinfo{person}{Yongchao Liu}, \bibinfo{person}{Xiaohe Bo}, \bibinfo{person}{Haizhou Shi}, \bibinfo{person}{Chuntao Hong}, \bibinfo{person}{Yan Zhang}, {and} \bibinfo{person}{Siliang Tang}.} \bibinfo{year}{2024}\natexlab{}.
\newblock \showarticletitle{Graph retrieval-augmented generation: A survey}.
\newblock \bibinfo{journal}{\emph{arXiv preprint arXiv:2408.08921}} (\bibinfo{year}{2024}).
\newblock


\bibitem[Petroni et~al\mbox{.}(2021)]%
        {petroni2021kilt}
\bibfield{author}{\bibinfo{person}{Fabio Petroni}, \bibinfo{person}{Aleksandra Piktus}, \bibinfo{person}{Angela Fan}, \bibinfo{person}{Patrick Lewis}, \bibinfo{person}{Majid Yazdani}, \bibinfo{person}{Nicola De~Cao}, \bibinfo{person}{James Thorne}, \bibinfo{person}{Yacine Jernite}, \bibinfo{person}{Vladimir Karpukhin}, \bibinfo{person}{Jean Maillard}, {et~al\mbox{.}}} \bibinfo{year}{2021}\natexlab{}.
\newblock \showarticletitle{KILT: a Benchmark for Knowledge Intensive Language Tasks}. In \bibinfo{booktitle}{\emph{Proceedings of the 2021 Conference of the North American Chapter of the Association for Computational Linguistics: Human Language Technologies}}. \bibinfo{pages}{2523--2544}.
\newblock


\bibitem[Sarthi et~al\mbox{.}({[n.\,d.]})]%
        {sarthi2024raptor}
\bibfield{author}{\bibinfo{person}{Parth Sarthi}, \bibinfo{person}{Salman Abdullah}, \bibinfo{person}{Aditi Tuli}, \bibinfo{person}{Shubh Khanna}, \bibinfo{person}{Anna Goldie}, {and} \bibinfo{person}{Christopher~D Manning}.} \bibinfo{year}{[n.\,d.]}\natexlab{}.
\newblock \showarticletitle{Raptor: Recursive abstractive processing for tree-organized retrieval}. In \bibinfo{booktitle}{\emph{The Twelfth International Conference on Learning Representations}}.
\newblock


\bibitem[Sonawane and Kulkarni(2014)]%
        {sonawane2014graph}
\bibfield{author}{\bibinfo{person}{Sheetal~S Sonawane} {and} \bibinfo{person}{Parag~A Kulkarni}.} \bibinfo{year}{2014}\natexlab{}.
\newblock \showarticletitle{Graph based representation and analysis of text document: A survey of techniques}.
\newblock \bibinfo{journal}{\emph{International Journal of computer applications}} \bibinfo{volume}{96}, \bibinfo{number}{19} (\bibinfo{year}{2014}), \bibinfo{pages}{1--8}.
\newblock


\bibitem[Song et~al\mbox{.}(2025)]%
        {song2025r1searcher}
\bibfield{author}{\bibinfo{person}{Huatong Song}, \bibinfo{person}{Jinhao Jiang}, \bibinfo{person}{Yingqian Min}, \bibinfo{person}{Jie Chen}, \bibinfo{person}{Zhipeng Chen}, \bibinfo{person}{Wayne~Xin Zhao}, \bibinfo{person}{Lei Fang}, {and} \bibinfo{person}{Ji-Rong Wen}.} \bibinfo{year}{2025}\natexlab{}.
\newblock \showarticletitle{R1-searcher: Incentivizing the search capability in llms via reinforcement learning}.
\newblock \bibinfo{journal}{\emph{arXiv preprint arXiv:2503.05592}} (\bibinfo{year}{2025}).
\newblock


\bibitem[Sui et~al\mbox{.}(2025)]%
        {sui2025stop}
\bibfield{author}{\bibinfo{person}{Yang Sui}, \bibinfo{person}{Yu-Neng Chuang}, \bibinfo{person}{Guanchu Wang}, \bibinfo{person}{Jiamu Zhang}, \bibinfo{person}{Tianyi Zhang}, \bibinfo{person}{Jiayi Yuan}, \bibinfo{person}{Hongyi Liu}, \bibinfo{person}{Andrew Wen}, \bibinfo{person}{Shaochen Zhong}, \bibinfo{person}{Hanjie Chen}, {et~al\mbox{.}}} \bibinfo{year}{2025}\natexlab{}.
\newblock \showarticletitle{Stop overthinking: A survey on efficient reasoning for large language models}.
\newblock \bibinfo{journal}{\emph{arXiv preprint arXiv:2503.16419}} (\bibinfo{year}{2025}).
\newblock


\bibitem[Sun et~al\mbox{.}({[n.\,d.]})]%
        {tog}
\bibfield{author}{\bibinfo{person}{Jiashuo Sun}, \bibinfo{person}{Chengjin Xu}, \bibinfo{person}{Lumingyuan Tang}, \bibinfo{person}{Saizhuo Wang}, \bibinfo{person}{Chen Lin}, \bibinfo{person}{Yeyun Gong}, \bibinfo{person}{Lionel Ni}, \bibinfo{person}{Heung-Yeung Shum}, {and} \bibinfo{person}{Jian Guo}.} \bibinfo{year}{[n.\,d.]}\natexlab{}.
\newblock \showarticletitle{Think-on-Graph: Deep and Responsible Reasoning of Large Language Model on Knowledge Graph}. In \bibinfo{booktitle}{\emph{The Twelfth International Conference on Learning Representations}}.
\newblock


\bibitem[Talmor et~al\mbox{.}(2019)]%
        {talmor2019commonsenseqa}
\bibfield{author}{\bibinfo{person}{Alon Talmor}, \bibinfo{person}{Jonathan Herzig}, \bibinfo{person}{Nicholas Lourie}, {and} \bibinfo{person}{Jonathan Berant}.} \bibinfo{year}{2019}\natexlab{}.
\newblock \showarticletitle{CommonsenseQA: A Question Answering Challenge Targeting Commonsense Knowledge}. In \bibinfo{booktitle}{\emph{Proceedings of the 2019 Conference of the North American Chapter of the Association for Computational Linguistics: Human Language Technologies, Volume 1 (Long and Short Papers)}}. \bibinfo{pages}{4149--4158}.
\newblock


\bibitem[Team et~al\mbox{.}(2025)]%
        {team2025kimi}
\bibfield{author}{\bibinfo{person}{Kimi Team}, \bibinfo{person}{Angang Du}, \bibinfo{person}{Bofei Gao}, \bibinfo{person}{Bowei Xing}, \bibinfo{person}{Changjiu Jiang}, \bibinfo{person}{Cheng Chen}, \bibinfo{person}{Cheng Li}, \bibinfo{person}{Chenjun Xiao}, \bibinfo{person}{Chenzhuang Du}, \bibinfo{person}{Chonghua Liao}, {et~al\mbox{.}}} \bibinfo{year}{2025}\natexlab{}.
\newblock \showarticletitle{Kimi k1. 5: Scaling reinforcement learning with llms}.
\newblock \bibinfo{journal}{\emph{arXiv preprint arXiv:2501.12599}} (\bibinfo{year}{2025}).
\newblock


\bibitem[Thakur et~al\mbox{.}({[n.\,d.]})]%
        {thakurbeir}
\bibfield{author}{\bibinfo{person}{Nandan Thakur}, \bibinfo{person}{Nils Reimers}, \bibinfo{person}{Andreas R{\"u}ckl{\'e}}, \bibinfo{person}{Abhishek Srivastava}, {and} \bibinfo{person}{Iryna Gurevych}.} \bibinfo{year}{[n.\,d.]}\natexlab{}.
\newblock \showarticletitle{BEIR: A Heterogeneous Benchmark for Zero-shot Evaluation of Information Retrieval Models}.
\newblock  (\bibinfo{year}{[n.\,d.]}).
\newblock


\bibitem[Trivedi et~al\mbox{.}(2022)]%
        {trivedi2021musique}
\bibfield{author}{\bibinfo{person}{Harsh Trivedi}, \bibinfo{person}{Niranjan Balasubramanian}, \bibinfo{person}{Tushar Khot}, {and} \bibinfo{person}{Ashish Sabharwal}.} \bibinfo{year}{2022}\natexlab{}.
\newblock \showarticletitle{{M}u{S}i{Q}ue: Multihop Questions via Single-hop Question Composition}.
\newblock \bibinfo{journal}{\emph{Transactions of the Association for Computational Linguistics}} (\bibinfo{year}{2022}).
\newblock


\bibitem[Vaswani et~al\mbox{.}(2017)]%
        {vaswani2017attention}
\bibfield{author}{\bibinfo{person}{Ashish Vaswani}, \bibinfo{person}{Noam Shazeer}, \bibinfo{person}{Niki Parmar}, \bibinfo{person}{Jakob Uszkoreit}, \bibinfo{person}{Llion Jones}, \bibinfo{person}{Aidan~N Gomez}, \bibinfo{person}{{\L}ukasz Kaiser}, {and} \bibinfo{person}{Illia Polosukhin}.} \bibinfo{year}{2017}\natexlab{}.
\newblock \showarticletitle{Attention is all you need}.
\newblock \bibinfo{journal}{\emph{Advances in neural information processing systems}}  \bibinfo{volume}{30} (\bibinfo{year}{2017}).
\newblock


\bibitem[Wang(2025)]%
        {wang2025proprag}
\bibfield{author}{\bibinfo{person}{Jingjin Wang}.} \bibinfo{year}{2025}\natexlab{}.
\newblock \showarticletitle{PropRAG: Guiding Retrieval with Beam Search over Proposition Paths}.
\newblock \bibinfo{journal}{\emph{arXiv preprint arXiv:2504.18070}} (\bibinfo{year}{2025}).
\newblock


\bibitem[Wang et~al\mbox{.}(2021)]%
        {wang2021ernie}
\bibfield{author}{\bibinfo{person}{Shuohuan Wang}, \bibinfo{person}{Yu Sun}, \bibinfo{person}{Yang Xiang}, \bibinfo{person}{Zhihua Wu}, \bibinfo{person}{Siyu Ding}, \bibinfo{person}{Weibao Gong}, \bibinfo{person}{Shikun Feng}, \bibinfo{person}{Junyuan Shang}, \bibinfo{person}{Yanbin Zhao}, \bibinfo{person}{Chao Pang}, {et~al\mbox{.}}} \bibinfo{year}{2021}\natexlab{}.
\newblock \showarticletitle{Ernie 3.0 titan: Exploring larger-scale knowledge enhanced pre-training for language understanding and generation}.
\newblock \bibinfo{journal}{\emph{arXiv preprint arXiv:2112.12731}} (\bibinfo{year}{2021}).
\newblock


\bibitem[WANG et~al\mbox{.}(2024)]%
        {wang2024current}
\bibfield{author}{\bibinfo{person}{Yaozu WANG}, \bibinfo{person}{Qing LI}, \bibinfo{person}{Zhangjie DAI}, {and} \bibinfo{person}{Yue XU}.} \bibinfo{year}{2024}\natexlab{}.
\newblock \showarticletitle{Current status and trends in large language modeling research}.
\newblock \bibinfo{journal}{\emph{Chinese Journal of Engineering}} \bibinfo{volume}{46}, \bibinfo{number}{8} (\bibinfo{year}{2024}), \bibinfo{pages}{1411--1425}.
\newblock


\bibitem[Wang et~al\mbox{.}(2024)]%
        {wang2024kgp}
\bibfield{author}{\bibinfo{person}{Yu Wang}, \bibinfo{person}{Nedim Lipka}, \bibinfo{person}{Ryan~A Rossi}, \bibinfo{person}{Alexa Siu}, \bibinfo{person}{Ruiyi Zhang}, {and} \bibinfo{person}{Tyler Derr}.} \bibinfo{year}{2024}\natexlab{}.
\newblock \showarticletitle{Knowledge graph prompting for multi-document question answering}. In \bibinfo{booktitle}{\emph{Proceedings of the AAAI conference on artificial intelligence}}, Vol.~\bibinfo{volume}{38}. \bibinfo{pages}{19206--19214}.
\newblock


\bibitem[Xiang et~al\mbox{.}(2025)]%
        {xiang2025use}
\bibfield{author}{\bibinfo{person}{Zhishang Xiang}, \bibinfo{person}{Chuanjie Wu}, \bibinfo{person}{Qinggang Zhang}, \bibinfo{person}{Shengyuan Chen}, \bibinfo{person}{Zijin Hong}, \bibinfo{person}{Xiao Huang}, {and} \bibinfo{person}{Jinsong Su}.} \bibinfo{year}{2025}\natexlab{}.
\newblock \showarticletitle{When to use graphs in rag: A comprehensive analysis for graph retrieval-augmented generation}.
\newblock \bibinfo{journal}{\emph{arXiv preprint arXiv:2506.05690}} (\bibinfo{year}{2025}).
\newblock


\bibitem[Xie et~al\mbox{.}(2025)]%
        {xie2025logic}
\bibfield{author}{\bibinfo{person}{Tian Xie}, \bibinfo{person}{Zitian Gao}, \bibinfo{person}{Qingnan Ren}, \bibinfo{person}{Haoming Luo}, \bibinfo{person}{Yuqian Hong}, \bibinfo{person}{Bryan Dai}, \bibinfo{person}{Joey Zhou}, \bibinfo{person}{Kai Qiu}, \bibinfo{person}{Zhirong Wu}, {and} \bibinfo{person}{Chong Luo}.} \bibinfo{year}{2025}\natexlab{}.
\newblock \showarticletitle{Logic-rl: Unleashing llm reasoning with rule-based reinforcement learning}.
\newblock \bibinfo{journal}{\emph{arXiv preprint arXiv:2502.14768}} (\bibinfo{year}{2025}).
\newblock


\bibitem[Xiong et~al\mbox{.}(2025)]%
        {xiong2025mcts}
\bibfield{author}{\bibinfo{person}{Guanming Xiong}, \bibinfo{person}{Haochen Li}, {and} \bibinfo{person}{Wen Zhao}.} \bibinfo{year}{2025}\natexlab{}.
\newblock \showarticletitle{MCTS-KBQA: Monte Carlo Tree Search for Knowledge Base Question Answering}.
\newblock \bibinfo{journal}{\emph{arXiv preprint arXiv:2502.13428}} (\bibinfo{year}{2025}).
\newblock


\bibitem[Yang et~al\mbox{.}(2024b)]%
        {yang2024kg}
\bibfield{author}{\bibinfo{person}{Rui Yang}, \bibinfo{person}{Haoran Liu}, \bibinfo{person}{Edison Marrese-Taylor}, \bibinfo{person}{Qingcheng Zeng}, \bibinfo{person}{Yuhe Ke}, \bibinfo{person}{Wanxin Li}, \bibinfo{person}{Lechao Cheng}, \bibinfo{person}{Qingyu Chen}, \bibinfo{person}{James Caverlee}, \bibinfo{person}{Yutaka Matsuo}, {et~al\mbox{.}}} \bibinfo{year}{2024}\natexlab{b}.
\newblock \showarticletitle{KG-Rank: Enhancing Large Language Models for Medical QA with Knowledge Graphs and Ranking Techniques}. In \bibinfo{booktitle}{\emph{Proceedings of the 23rd Workshop on Biomedical Natural Language Processing}}. \bibinfo{pages}{155--166}.
\newblock


\bibitem[Yang et~al\mbox{.}(2024a)]%
        {yang2024dollmreasoning}
\bibfield{author}{\bibinfo{person}{Sohee Yang}, \bibinfo{person}{Elena Gribovskaya}, \bibinfo{person}{Nora Kassner}, \bibinfo{person}{Mor Geva}, {and} \bibinfo{person}{Sebastian Riedel}.} \bibinfo{year}{2024}\natexlab{a}.
\newblock \showarticletitle{Do Large Language Models Latently Perform Multi-Hop Reasoning?}. In \bibinfo{booktitle}{\emph{Proceedings of the 62nd Annual Meeting of the Association for Computational Linguistics (Volume 1: Long Papers)}}. \bibinfo{pages}{10210--10229}.
\newblock


\bibitem[Yang et~al\mbox{.}(2018)]%
        {yang2018hotpotqa}
\bibfield{author}{\bibinfo{person}{Zhilin Yang}, \bibinfo{person}{Peng Qi}, \bibinfo{person}{Saizheng Zhang}, \bibinfo{person}{Yoshua Bengio}, \bibinfo{person}{William~W. Cohen}, \bibinfo{person}{Ruslan Salakhutdinov}, {and} \bibinfo{person}{Christopher~D. Manning}.} \bibinfo{year}{2018}\natexlab{}.
\newblock \showarticletitle{{HotpotQA}: A Dataset for Diverse, Explainable Multi-hop Question Answering}. In \bibinfo{booktitle}{\emph{Conference on Empirical Methods in Natural Language Processing ({EMNLP})}}.
\newblock


\bibitem[Zeng et~al\mbox{.}(2024)]%
        {zeng2024large}
\bibfield{author}{\bibinfo{person}{Jingying Zeng}, \bibinfo{person}{Richard Huang}, \bibinfo{person}{Waleed Malik}, \bibinfo{person}{Langxuan Yin}, \bibinfo{person}{Bojan Babic}, \bibinfo{person}{Danny Shacham}, \bibinfo{person}{Xiao Yan}, \bibinfo{person}{Jaewon Yang}, {and} \bibinfo{person}{Qi He}.} \bibinfo{year}{2024}\natexlab{}.
\newblock \showarticletitle{Large language models for social networks: Applications, challenges, and solutions}.
\newblock \bibinfo{journal}{\emph{arXiv preprint arXiv:2401.02575}} (\bibinfo{year}{2024}).
\newblock


\bibitem[Zhang et~al\mbox{.}(2025b)]%
        {zhang2025corag}
\bibfield{author}{\bibinfo{person}{Jingyan Zhang}, \bibinfo{person}{Dawei Feng}, {and} \bibinfo{person}{Bo Ding}.} \bibinfo{year}{2025}\natexlab{b}.
\newblock \showarticletitle{CoRAG: Enhancing Retrieval-Augmented Generation with Coreference Resolution}. In \bibinfo{booktitle}{\emph{2025 5th International Conference on Artificial Intelligence and Industrial Technology Applications (AIITA)}}. IEEE, \bibinfo{pages}{2116--2123}.
\newblock


\bibitem[Zhang et~al\mbox{.}(2025a)]%
        {zhangsurvey}
\bibfield{author}{\bibinfo{person}{Qinggang Zhang}, \bibinfo{person}{Shengyuan Chen}, \bibinfo{person}{Yuanchen Bei}, \bibinfo{person}{Zheng Yuan}, \bibinfo{person}{Huachi Zhou}, \bibinfo{person}{Zijin Hong}, \bibinfo{person}{Junnan Dong}, \bibinfo{person}{Hao Chen}, \bibinfo{person}{Yi Chang}, {and} \bibinfo{person}{Xiao Huang}.} \bibinfo{year}{2025}\natexlab{a}.
\newblock \showarticletitle{A survey of graph retrieval-augmented generation for customized large language models}.
\newblock \bibinfo{journal}{\emph{arXiv preprint arXiv:2501.13958}} (\bibinfo{year}{2025}).
\newblock


\bibitem[Zhu et~al\mbox{.}(2024)]%
        {zhu2024structugraphrag}
\bibfield{author}{\bibinfo{person}{Xishi Zhu}, \bibinfo{person}{Xiaoming Guo}, \bibinfo{person}{Shengting Cao}, \bibinfo{person}{Shenglin Li}, {and} \bibinfo{person}{Jiaqi Gong}.} \bibinfo{year}{2024}\natexlab{}.
\newblock \showarticletitle{StructuGraphRAG: Structured document-informed knowledge graphs for retrieval-augmented generation}. In \bibinfo{booktitle}{\emph{Proceedings of the AAAI Symposium Series}}, Vol.~\bibinfo{volume}{4}. \bibinfo{pages}{242--251}.
\newblock


\end{thebibliography}

\appendix

\begin{table*}
  \caption{The results of different methods for multi-hop question answering using Qwen2.5-7B-Instruct. The first three datasets are in-domain, and PopQA is out-of-domain (i.e., unseen during training). The best results are in \textbf{bold} and the second-best results are \underline{underlined}.}
  \label{tab:v2_main}
\begin{tabular}{cccccccccccccc}
    \toprule
       \multirow{2}{*}{Method}    & \multicolumn{3}{c} {HotpotQA}    & \multicolumn{3}{c}{MuSiQue}    & \multicolumn{3}{c}{2Wiki}  & \multicolumn{3}{c}{PopQA}\\

              \cmidrule[1pt](lr){2-4}\cmidrule[1pt](lr){5-7}\cmidrule[1pt](lr){8-10}\cmidrule[1pt](lr){11-13}\
                            & $\mathrm{F1}$                          & $\mathrm{ACC_{L}}$               & $\mathrm{SBERT}$       & $\mathrm{F1}$                           & $\mathrm{ACC_{L}}$                 & $\mathrm{SBERT}$     & $\mathrm{F1}$                          & $\mathrm{ACC_{L}}$                   & $\mathrm{SBERT}$   & $\mathrm{F1}$  & $\mathrm{ACC_{L}}$                  & $\mathrm{SBERT}$\\
    \midrule

          \text{Naive RAG}      &  73.44    & \underline{79.50}   &   86.41  &   31.61 & 33.00  &  65.67  &   41.53  &  42.00   &  69.69   &    \underline{51.54}   &  \underline{62.90} & 73.92 \\
       \text{KGP}              & 45.90  &  51.00   &   70.56  &   10.09  &  13.00   &   50.04  &  17.73  &  22.50  & 55.56   &   42.39  &    54.50   &  69.21   \\
            \text{ToG}         & 20.41  &   25.00  &  56.73   &  10.30   &   12.00  &   49.66  &  29.13  &   29.50  & 62.16  &   30.67  &   35.00    &  63.46   \\

      \text{LightRAG}         &  42.74 &    47.00 &  70.84   &  18.92   &  16.50   & 57.84    &  29.59  & 30.00    &  62.72 &    47.65 &   59.00    &  71.73   \\

        \text{PropRAG}    & 71.11  &  77.50   &  85.71   &  38.88   &   \underline{42.00}  &   \underline{69.40}  &  \underline{60.31} &   \underline{63.00}  &  \underline{79.45} &  44.43  &   58.30    &  69.16   \\
        \text{G-retriever}    & 15.74  &  19.50   &  55.52   &  6.19   &   6.00  &    48.49 &   12.80 &   52.20  &  12.00 &   2.52  &   4.10    &  43.82   \\

        \text{HippoRAG2}      &  \underline{73.91} &   78.00  &   \underline{86.44}  &   \underline{40.16}  &  39.00   & 69.27    &    51.28  &  54.00   & 75.63   &  51.27   &   \textbf{63.00}    &   \underline{74.05}  \\
     \midrule
 \text{\methodname}     &  \textbf{74.21} &  \textbf{80.00}    &  \textbf{87.40}   &  \textbf{49.63}   &  \textbf{50.00}   &   \textbf{74.27}  &   \textbf{68.28} &   \textbf{74.00}  &  \textbf{83.57} & \textbf{57.91}    & 61.50      & \textbf{79.75}    \\

    \bottomrule
  \end{tabular}
\end{table*}

\begin{table*}
  \setlength{\tabcolsep}{3.2pt}
  \caption{The results of integrating \methodname~using Qwen2.5-7B-Instruct~ with other GraphRAG retrieval methods. Imp denotes the average improvement across all datasets.}
  \label{tab:qwen_ins_integrate}
  \begin{tabular}{clccccccccccccc}
    \toprule

    \multirow{2}{*}{Method} & \multirow{2}{*}{Model}      & \multicolumn{3}{c} {HotpotQA}    & \multicolumn{3}{c}{MuSiQue}    & \multicolumn{3}{c}{2Wiki}  & \multicolumn{3}{c}{PopQA}   & \multirow{2}{*}{Imp}\\
              \cmidrule[1pt](lr){3-5}\cmidrule[1pt](lr){6-8}\cmidrule[1pt](lr){9-11}\cmidrule[1pt](lr){12-14}\
                         &    & $\mathrm{F1}$                          & $\mathrm{ACC_{L}}$               & $\mathrm{SBERT}$       & $\mathrm{F1}$                           & $\mathrm{ACC_{L}}$                 & $\mathrm{SBERT}$     & $\mathrm{F1}$                          & $\mathrm{ACC_{L}}$                   & $\mathrm{SBERT}$   & $\mathrm{F1}$  & $\mathrm{ACC_{L}}$                  & $\mathrm{SBERT}$\\
   \midrule

       \multirow{2}{*}{Naive RAG}& \text{Original}  &  \textbf{73.44}    & 79.50   &   \textbf{86.41}  &   31.61 & 33.00  &  65.67  &   41.53  &  42.00   &  69.69   &    \textbf{51.54}   &  \textbf{62.90} & \textbf{73.92} \\
   &   \text{+\methodname} &    71.51   &  \textbf{80.00}  &  85.30   &     \textbf{51.96} &   \textbf{53.50}    &    \textbf{75.38}   &   \textbf{56.77}   &   \textbf{67.00}  &   \textbf{76.06}    &  42.29    & 59.60 &68.55 & +10.79\%\\
    \midrule
    
    \multirow{2}{*}{KGP} 
        & \text{Original}     
        & 45.90  &  51.00   &   70.56  &   10.09  &  13.00   &   50.04  &  17.73  &  22.50  & 55.56   &   \textbf{42.39}  &    \textbf{54.50}   &  \textbf{69.21}   \\
        & \text{+\methodname} 
        & \textbf{52.27}    & \textbf{60.50}    & \textbf{75.66}     & \textbf{27.26}     & \textbf{27.50}        & \textbf{62.76}         & \textbf{44.83}        & \textbf{55.00}                & \textbf{70.92}         & 33.99                  & 49.70         & 64.46 & +24.35\% \\
    \midrule

    \multirow{2}{*}{ToG} 
        & \text{Original}     
         & 20.41  &   25.00  &  56.73   &  10.30   &   12.00  &   49.66  &  29.13  &   29.50  & 62.16  &   \textbf{30.67}  &   35.00    &  \textbf{63.46}   \\
        & \text{+\methodname} 
        & \textbf{21.44}        & \textbf{27.00}          & \textbf{61.63}         & \textbf{21.99}        & \textbf{21.50}                  & \textbf{59.28}         & \textbf{44.96}        & \textbf{54.50}                & \textbf{70.61}         & 27.24                  & \textbf{37.20}         & 60.83 & +19.85\% \\
    \midrule

    \multirow{2}{*}{LightRAG} 
        & \text{Original}     
        &  42.74 &    47.00 &  70.84   &  18.92   &  16.50   & 57.84    &  29.59  & 30.00    &  62.72 &    \textbf{47.65} &   \textbf{59.00}    &  \textbf{71.73}   \\
        & \text{+\methodname} 
        & \textbf{56.12}         & \textbf{63.00} & \textbf{77.40}         & \textbf{40.10}        & \textbf{40.00}         & \textbf{68.21}         & \textbf{55.99}        & \textbf{63.40}       & \textbf{76.13}         & 42.74                  & 57.10        & 69.57 & +27.99\% \\
    \midrule
  
       \multirow{2}{*}{PropRAG} 
        & \text{Original}     
        & \textbf{71.11}  &  77.50   &  \textbf{85.71}   &  38.88   &   42.00  &   69.40  &  60.31 &   63.00  &  79.45 &  \textbf{44.43}  &   58.30    &  69.16   \\

        & \text{+\methodname} 
        & 69.81                & \textbf{78.00}         & 83.44         & \textbf{49.94}        & \textbf{52.50}         & \textbf{74.97}         & \textbf{68.23}         & \textbf{77.50}       & \textbf{82.51}         & 44.22                 & \textbf{60.30}         & \textbf{69.93} & +6.86\% \\
        \midrule

\multirow{2}{*}{G-retriever} 
& \text{Original}     
 & 15.74  &  19.50   &  55.52   &  6.19   &   6.00  &    48.49 &   12.80 &   52.20  &  12.00 &   2.52  &   4.10    &  43.82   \\
& \text{+\methodname} 
& \textbf{31.47}        & \textbf{30.50} & \textbf{66.45}         & \textbf{16.13}        & 12.00                 & \textbf{56.87}         & \textbf{22.96}        & \textbf{22.50}       & \textbf{59.56}         & \textbf{15.74}         & \textbf{16.90}        & \textbf{53.92} & +45.22\% \\
\midrule

\multirow{2}{*}{HippoRAG2} 
& \text{Original}     
&  73.91 &   78.00  &   86.44  &   40.16  &  39.00   & 69.27    &    51.28  &  54.00   & 75.63   &  51.27   &   \textbf{63.00}    &   74.05  \\
& \text{+\methodname} 
&  \textbf{74.21} &  \textbf{80.00}    &  \textbf{87.40}   &  \textbf{49.63}   &  \textbf{50.00}   &   \textbf{74.27}  &   \textbf{68.28} &   \textbf{74.00}  &  \textbf{83.57} & \textbf{57.91}    & 61.50      & \textbf{79.75}& +9.87\%     \\
    \bottomrule
  \end{tabular}
\end{table*}

\section{More Experiments}

\subsection{Experiments with Qwen2.5-7B-Instruct}
In this study, we further explore the performance of \methodname~on instruct-tuned models. We observed that results based on the original Qwen2.5-7B-base still have room for improvement on certain tasks, particularly in terms of instruction compliance and contextual understanding. Therefore, we selected the Qwen2.5-7B-Instruct model as the backbone. Compared to the base model, the instruct-tuned model demonstrates stronger capabilities in understanding user intent, following output formats, and handling multi-turn dialogues. Based on this model, we retained the original GraphRAG-R1 framework and only performed adaptive tuning on the training strategy, prompt templates, and inference hyperparameters, aiming to verify the stability and scalability of our method across different types of backbone models.

The experimental results demonstrate that the tuned \methodname~with Qwen2.5-7B-Instruct not only significantly outperforms the base version but also achieves the best performance on most evaluation metrics across datasets such as HotpotQA, MuSiQue, 2Wiki, and PopQA. Detailed performance comparisons are provided in Table~\ref{tab:v2_main}.

Furthermore, we tested the integration of our approach with other retrievers, as shown in Table~\ref{tab:qwen_ins_integrate}. In all tested configurations, \methodname~consistently improved performance over the original retrieval methods, demonstrating strong transferability and generalization. This indicates that our framework can be flexibly adapted to different retrieval mechanisms.

\subsection{Experiments with LLaMA-3-8B}

\begin{table*}
  \caption{The results of different methods for multi-hop question answering using Llama-3-8B as the backbone LLM. %
  The best results are in \textbf{bold} and the second-best results are \underline{underlined}.}
  \label{tab:llamamain}
  \begin{tabular}{cccccccccccccc}
    \toprule
       \multirow{2}{*}{Method}    & \multicolumn{3}{c} {HotpotQA}    & \multicolumn{3}{c}{MuSiQue}    & \multicolumn{3}{c}{2Wiki}  & \multicolumn{3}{c}{PopQA}\\

              \cmidrule[1pt](lr){2-4}\cmidrule[1pt](lr){5-7}\cmidrule[1pt](lr){8-10}\cmidrule[1pt](lr){11-13}\
                            & $\mathrm{F1}$                          & $\mathrm{ACC_{L}}$               & $\mathrm{SBERT}$       & $\mathrm{F1}$                           & $\mathrm{ACC_{L}}$                 & $\mathrm{SBERT}$     & $\mathrm{F1}$                          & $\mathrm{ACC_{L}}$                   & $\mathrm{SBERT}$   & $\mathrm{F1}$  & $\mathrm{ACC_{L}}$                  & $\mathrm{SBERT}$\\
    \midrule

          \text{Naive RAG}      &  9.57 &  \textbf{27.00}  & 48.25   &  1.17  &  4.00  &  40.11   &  4.14  &  10.00  & 43.58  &  8.14  &      24.70   &   46.59  \\
       \text{KGP}              &    3.61   &  17.00 &  46.41    &   1.38  &    3.50  &  44.49   &  \underline{8.23}   &     \textbf{26.50}    &    \underline{48.48}   &    10.75  &  \textbf{59.00}  &     48.13    \\
            \text{ToG}         & 8.97  &  16.50 &  \underline{49.83}  &    3.28  & 10.00     &  \underline{44.74}  &   7.91 &   23.00  &  48.36  &  \underline{10.96}  &   37.50   &  \underline{49.57}      \\

      \text{LightRAG}         &  5.33 &  12.50 &   43.95 &  1.83   &   3.50    &  41.51   &   4.68  &   12.00 &  43.95  & 6.71 &   27.50 & 45.83       \\

        \text{PropRAG}    & \underline{12.47} &  23.00 &  49.03  &  \underline{3.83}   &   6.50    &  42.17   &   5.06   &  11.50  & 43.06  &  9.76 &   25.70   &48.10      \\
        \text{G-retriever}    & 1.91  &  23.00 &   43.09 &  1.06   &  \underline{12.50}    &  42.77   &   2.43  &   18.00   &  42.26  &   0.25 & 38.10     &  39.32  \\

        \text{HippoRAG2}       & 6.89 &  22.00  &   46.06  &   1.85 &   4.00   &   42.03  & 5.61 &   12.00  &  43.51 &  7.85 &   26.10   &     46.57 \\
     \midrule
 \text{\methodname} & \textbf{13.16}  &  \underline{26.50} &   \textbf{53.06}  &  \textbf{6.32}  &    \textbf{16.00}    &  \textbf{48.96}   &    \textbf{13.36}  &   \underline{25.00}  &    \textbf{51.46} &    \textbf{11.64}  &  \underline{41.50}    &     \textbf{50.42}    \\

    \bottomrule
  \end{tabular}
\end{table*}

In this subsection, we investigate the effectiveness of \methodname~ across different base models. We adopt another representative open-sourced LLM, Llama-3-8B, as the backbone LLM. The experimental setting is similar to using Qwen2.5-7B, except we add PRA reward in the third training stage to encourage the model to invoke retrieval, which enhances performance. 
The overall results are presented in Table~\ref{tab:llamamain}.
Collectively, these experiments demonstrate that our method is effective and outperforms baselines when using LLaMA-3-8B as the backbone LLM, showing that our method is generalizable across different LLM models.

\subsection{Hyperparameter Sensitivity Analysis}

\begin{figure}[ht]
  \includegraphics[width=\linewidth,page=1]{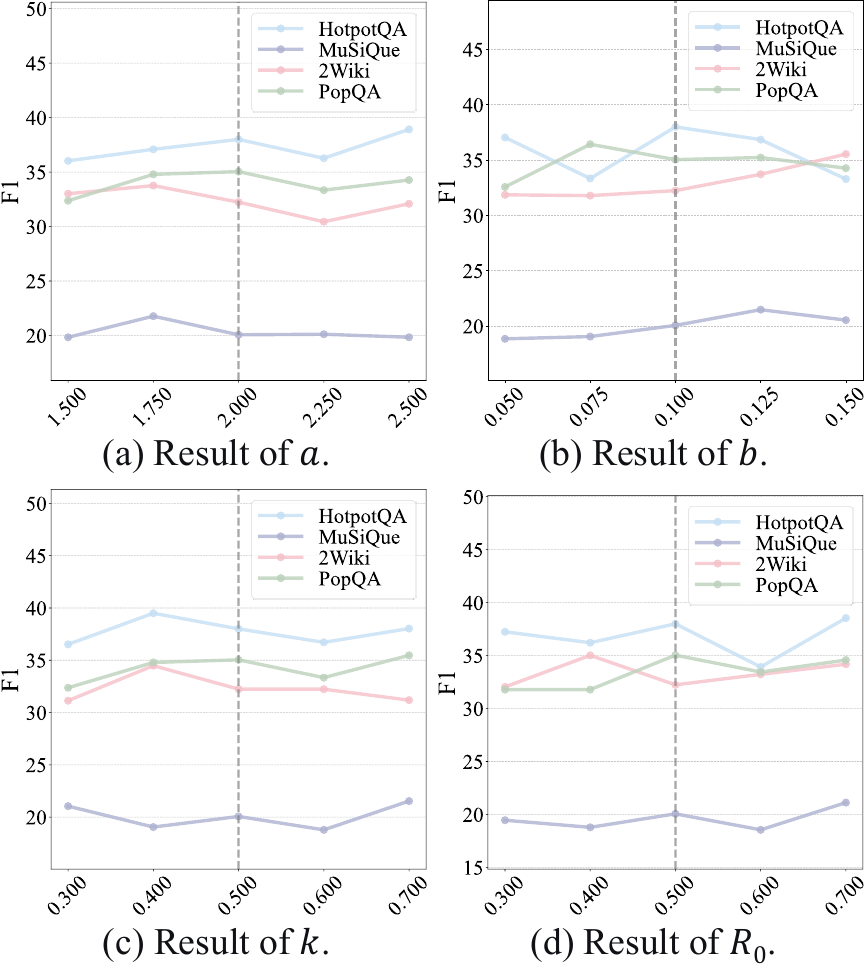}
  \vspace{-0.3cm}
  \caption{Sensitivity analysis of key hyperparameters in \methodname. Subfigures (a)-(d) illustrate the variation in F1 score on the four validation sets as the key hyperparameters are varied within specified ranges. Other hyperparameters were fixed at their main experiment values. The gray dashed lines indicate the parameter values selected for the main experiments. }
\Description{
Sensitivity analysis of key hyperparameters in GraphRAG-R1. Subfigures (a)-(d) illustrate the variation in F1 score on the four validation sets as the key hyperparameters are varied within specified ranges. Other hyperparameters were fixed at their main experiment values. The gray dashed lines indicate the parameter values selected for the main experiments.
}
  \label{fig:hyper}
  \vspace{-0.3cm}
\end{figure}

To evaluate the hyperparameter robustness of the \methodname~framework, we conducted sensitivity analysis on four key parameters. The results are shown in Figure~\ref{fig:hyper} and demonstrate that the performance of \methodname~has a wide robustness interval around the parameters selected in the main experiment, which enhances the reproducibility and practical reliability of the method.

\subsection{Cost and Efficiency Analysis}

\begin{table*}
  \caption{Performance and cost comparison with KGP and ToG, two reasoning-enhanced GraphRAG methods}
  \vspace{-0.3cm}
  \label{tab:togkgp}
  \begin{tabular}{ccccccccc}
    \toprule
    \multirow{2}{*}{Method}      & \multicolumn{2}{c}{HotpotQA}    & \multicolumn{2}{c}{MuSiQue}    & \multicolumn{2}{c}{2Wiki} & \multicolumn{2}{c}{PopQA}\\

              \cmidrule[1pt](lr){2-3}\cmidrule[1pt](lr){4-5}\cmidrule[1pt](lr){6-7}\cmidrule[1pt](lr){8-9}\
                             & $\mathrm{F1}$              & \#$\mathrm{Token}$                                           & $\mathrm{F1}$              & \#$\mathrm{Token}$                                            & $\mathrm{F1}$              & \#$\mathrm{Token}$              & $\mathrm{F1}$              & \#$\mathrm{Token}$         \\
    \midrule
    \text{KGP}  &   10.73  & 2687.44     &  4.61    & 2858.23     &   10.16&  2173.02          &    21.01    &   3182.02    \\
    \text{ToG}  &  11.44   &  3136.61   &  5.02   &   2873.56    &  14.45   &    2065.94&  29.21  & 2838.49    \\
    \text{\methodname}         &   \textbf{38.00}    &    \textbf{1674.47} &  \textbf{20.06}   &  \textbf{2086.10}  &   \textbf{32.24}   &  \textbf{1883.21}        &   \textbf{35.04}    &  \textbf{1449.80}    \\        
    \bottomrule
  \end{tabular}
\end{table*}

\begin{table*}
\caption{The results of integrating \methodname~with other GraphRAG retrieval methods. Note that \methodname~is only trained once and do not access other RAG methods during training. Imp denotes the average improvement across all datasets. }
\vspace{-0.3cm}
  \label{tab:dalk}
  \setlength{\tabcolsep}{3.5pt}
  \begin{tabular}{clccccccccccccc}
    \toprule
    \multirow{2}{*}{Method} & \multirow{2}{*}{Model}      & \multicolumn{3}{c} {HotpotQA}    & \multicolumn{3}{c}{MuSiQue}    & \multicolumn{3}{c}{2Wiki}  & \multicolumn{3}{c}{PopQA}   & \multirow{2}{*}{Imp}\\

              \cmidrule[1pt](lr){3-5}\cmidrule[1pt](lr){6-8}\cmidrule[1pt](lr){9-11}\cmidrule[1pt](lr){12-14}\
                         &    & $\mathrm{F1}$                          & $\mathrm{ACC_{L}}$               & $\mathrm{SBERT}$       & $\mathrm{F1}$                           & $\mathrm{ACC_{L}}$                 & $\mathrm{SBERT}$     & $\mathrm{F1}$                          & $\mathrm{ACC_{L}}$                   & $\mathrm{SBERT}$   & $\mathrm{F1}$  & $\mathrm{ACC_{L}}$                  & $\mathrm{SBERT}$  \\
  \midrule

    \multirow{2}{*}{Dalk}&   Original &  0.62  &  7.50  &   41.96  &  0.40  &    3.00   &  40.64 &   0.75  &    4.00  &  41.66 &   0.14   &    8.80     &    39.37     \\
    & +\methodname & 15.95  & 20.50   &    52.38  &  8.15  &   8.50   &  48.33  &  10.45   & 14.40        &   49.54  &    13.38  & 26.60 & 51.90& + 69.50\%\\

    \bottomrule
  \end{tabular}
\end{table*}

In this subsection, we evaluate the performance and cost of different reasoning-enhanced retrieval methods. We compare our approach with KGP and ToG, two reasoning-enhanced retrieval methods that also support multiple retrievals. In addition to comparing the performance measured by F1 scores, we also calculate the token consumption throughout the retrieval and reasoning processes. Specifically, for ToG and KGP, we count the number of accesses to the knowledge graph, for our method, we tally the frequency of calls to external retrieval tools. Token overhead accounts for consumption throughout the retrieval and reasoning processes.
As shown in Table ~\ref{tab:togkgp}, \methodname~ achieves significant improvements in both performance and computational cost by deeply integrating the intrinsic reasoning capabilities of LLMs. Compared to baseline methods that also leverage LLM reasoning, our method demonstrates substantially improved answer quality while requiring lower token overhead than the other two methods. This validates the effectiveness and cost efficiency of our approach.

\subsection{Cross-Domain Adaptation Verification}

\methodname~is designed as a general framework agnostic to specific graph structures, tasks, and domains. To verify its domain generalization and compatibility, we adapted an advanced retrieval method DALK~\cite{li2024dalk} originally designed for the medical domain, integrated it into our framework, and evaluated it on general-domain datasets.
We only replacing \methodname's underlying retrieval tool with the adapted DALK retriever. As shown in Table~\ref{tab:dalk}, the integrated framework achieved a significant improvement compared to the original DALK method, with a relative F1 score increase of approximately 69.50\% on Qwen2.5-7B. This strongly proves that our framework can flexibly compatibly integrate and enhance specialized retrievers from different domains.

\subsection{More Details about Datasets}\label{sec:dataset}
\begin{table}[htbp]
 \setlength{\tabcolsep}{2pt}
\centering
\caption{Dataset Statistics}
\vspace{-0.4cm}
\begin{tabular}{clcccc}
\toprule
    & Metric & MuSiQue & 2Wiki & HotpotQA & PopQA \\
\midrule
\multirow{3}{*}{Graph} &\# Passages & 11,656  & 6,119  & 9,221   & 8,676 \\
&\# Nodes & 91,729 & 42,694 &  82,157  & 74,472 \\
&\# Triplets & 107,448 & 50,671 &  98,709 & 139,273 \\
\midrule
\multirow{3}{*}{Questions} & Total & 1,000  & 1,000  & 1,000   & 1,000 \\
&Training Set & 800 & 800 &  800  & 0 \\
&Testing Set  & 200 & 200 &  200 & 1,000 \\
\bottomrule
\end{tabular}
\label{tab:qa_datasets_stats}
\end{table}

We adopt four widely used benchmark datasets, including HotpotQA, MuSiQue, and 2Wiki from HippoRAG~\cite{jimenez2024hipporag}\footnote{\url{https://github.com/OSU-NLP-Group/HippoRAG/releases/tag/v1.0.0}} and one simple QA dataset PopQA from PropRAG~\cite{wang2025proprag}\footnote{\url{https://github.com/ReLink-Inc/PropRAG/tree/main/outputs/popqa}}. Detailed descriptions of the datasets are as follows.
\begin{itemize}[leftmargin=0.5cm]
    \item \textbf{HotpotQA}~\cite{yang2018hotpotqa}: A Wikipedia-based dataset where answering each question requires reasoning over multiple supporting documents.
    \item \textbf{MuSiQue}~\cite{trivedi2021musique}: A challenging dataset consisting of questions composed of 2 to 4 single-hop questions, requiring multi-step reasoning.
    \item \textbf{2Wiki}~\cite{ho2020constructing}: Another Wikipedia-based dataset where each question requires information from two articles to be correctly answered.
    \item \textbf{PopQA}~\cite{mallen2023not}: A dataset of entity-centric QA pairs, where each question is generated by converting a knowledge tuple retrieved from Wikidata.
\end{itemize}
The statistics of these datasets are shown in Table \ref{tab:qa_datasets_stats}. %

\end{document}